\documentclass[11pt]{article}
\usepackage{amssymb}
\usepackage{amsmath}
\usepackage{amsthm}
\usepackage{amsbsy}
\usepackage{subfigure}
\usepackage{natbib}
\usepackage{fullpage}

\usepackage{booktabs}

\RequirePackage[colorlinks,citecolor=blue,urlcolor=blue,breaklinks]{hyperref}
 
\newtheorem{assumption}{Assumption}
\newtheorem{theorem}{Theorem}
\newtheorem{lemma}[theorem]{Lemma} 
 
\newtheorem{remark}{Remark}

\def\proclaim#1{\par \bigskip\noindent {\bf #1}\bgroup\it\ }
\def\endproclaim{\egroup\par\bigskip}
\def\proof#1{\par\noindent{\bf #1} \;}

\usepackage{bm}

\usepackage{graphicx}
\usepackage{appendix}

\usepackage{algorithm}
\usepackage{algpseudocode}

\usepackage{caption}  
\makeatletter
\newcommand\footnoteref[1]{\protected@xdef\@thefnmark{\ref{#1}}\@footnotemark}
\makeatother

\usepackage{xcolor}
\definecolor{DSgray}{cmyk}{0,1,0,0}

\begin{document}
\title{\bf A Bias-Correction Decentralized Stochastic Gradient Algorithm with Momentum Acceleration}

\author{Yuchen Hu\footnote{\scriptsize School of Mathematical Sciences, Shanghai Jiao Tong University, Shanghai, 200240, China, \texttt{huyuchen\_stat@sjtu.edu.cn}} \quad 
Xi Chen\footnote{\scriptsize New York University, New York, NY 10012, USA, \texttt{xc13@stern.nyu.edu}} \quad 
Weidong Liu\footnote{School of Mathematical Sciences, Shanghai Jiao Tong University, Shanghai, 200240, China, \texttt{\scriptsize weidongl@sjtu.edu.cn}}\quad 
Xiaojun Mao\footnote{\scriptsize School of Mathematical Sciences, Shanghai Jiao Tong University, Shanghai, 200240, China, \texttt{maoxj@sjtu.edu.cn}}}

\date{}
\maketitle
    
\begin{abstract}
Distributed stochastic optimization algorithms can process large-scale datasets simultaneously, significantly accelerating model training. However, their effectiveness is often hindered by the sparsity of distributed networks and data heterogeneity. In this paper, we propose a momentum-accelerated distributed stochastic gradient algorithm, termed \textbf{E}xact-\textbf{D}iffusion with \textbf{M}omentum (\textbf{EDM}), which mitigates the bias from data heterogeneity and incorporates momentum techniques commonly used in deep learning to enhance convergence rate. Our theoretical analysis demonstrates that the EDM algorithm converges sub-linearly to the neighborhood of the optimal solution, the radius of which is irrespective of data heterogeneity, when applied to non-convex objective functions; under the Polyak-\L ojasiewicz condition, which is a weaker assumption than strong convexity, it converges linearly to the target region. Our analysis techniques employed to handle momentum in complex distributed parameter update structures yield a sufficiently tight convergence upper bound, offering a new perspective for the theoretical analysis of other momentum-based distributed algorithms.
\end{abstract}

\section{Introduction}\label{sec:Intro}
The distributed stochastic optimization (DSO) problem has garnered significant attention across various domains, including the Internet of Things (IoT) \citep{ali2018applications}, federated learning \citep{beltran2023decentralized}, and statistical machine learning \citep{boyd2011distributed}. In this paper, we specifically address the following optimization problem:
\begin{equation}\label{eq: general problem}
    \mathop{\min}\limits_{\mathbf{x} \in \mathbb{R}^d}\frac1n \sum_{i=1}^n f_i(\mathbf{x}),\quad  f_i(\mathbf{x}) \triangleq \mathbb{E}_{\bm\xi_i \sim \mathcal{D}_i} F_i(\mathbf{x}, \bm\xi_i),
\end{equation} 
where $F_i(\mathbf{x},\bm\xi_i)$  represents the loss function at agent $i$ given the parameter $\mathbf{x}\in \mathbb{R}^d$, and $\bm\xi_i$ denotes the data on agent $i$, which is independently and identically generated from the distribution $\mathcal{D}_i$. Each agent is capable of exchanging information with its neighbors, thus facilitating collaborative decision-making to optimize the overall global objective function.

A common approach to solving DSO problems is through first-order methods \citep{xin2020general}. Notably, \cite{nedic2009distributed} proposed a sub-gradient optimization algorithm for distributed optimization (DO), which serves as the deterministic counterpart of DSO. By leveraging the concept of stochastic approximation, these algorithms can be effectively adapted to tackle distributed stochastic optimization problems by estimating the gradient corresponding to the current parameters of each agent. \cite{jiang2017collaborative} and \cite{lian2017can} introduced a stochastic variant of \cite{yuan2016convergence}, referred to as Distributed Stochastic Gradient Descent (DSGD), and investigated its convergence properties for non-convex objective functions. Following this, a series of improved stochastic optimization algorithms were developed \citep{spiridonoff2020robust,zhang2019decentralized,tang2018d}. 

In a distributed scenario, data heterogeneity among agents presents a significant challenge \citep{hsieh2020non}. Several works \cite{yuan2021decentlam,lin2021quasi} have indicated that DSGD can converge to a neighborhood of the optimal value, with the radius of this neighborhood determined by the variance of the stochastic gradient and the heterogeneity of the data for both strongly convex optimization and non-convex optimization. The sparsity of the network further exacerbates the impact of heterogeneity \citep{koloskova2020unified,yuan2020influence}. A viable strategy to mitigate this issue is to modify the gradient update to approximate the average gradient, collectively referred to as bias-correction algorithms. For instance, \cite{zhang2019decentralized} proposed Distributed Stochastic Gradient Tracking (DSGT), a widely used bias-correction algorithm, and demonstrated its sub-linear convergence for non-convex objective functions. Its convergence properties in the strongly convex case were established in \cite{pu2021distributed}. In contrast, exact-diffusion \citep{yuan2020influence}, also denoted as D$^2$ in \cite{tang2018d}, effectively eliminates the effects of data heterogeneity without requiring additional information aggregation. A unified analysis of the bias-correction algorithms is conducted in \cite{Unified_D2_yuan2022}, and \cite{alghunaim2024enhanced} noted that the convergence rate of DSGT is slightly inferior to ED/D$^2$.

To enhance convergence properties, a common strategy is to incorporate the momentum method \citep{nesterov2013introductory,polyak1964some}. Works by \cite{yu2019linear} and \cite{gao2020periodic} have introduced momentum into distributed stochastic optimization problems, providing convergence rates in the non-convex case. However, the introduction of momentum does not eliminate heterogeneous bias, which can render the convergence results less robust \cite{yuan2021decentlam,takezawa2023momentum}. Some methods, such as DecentLaM \cite{yuan2021decentlam} and Quasi-Global \cite{lin2021quasi}, partially address this issue by modifying the momentum architecture. Introducing momentum into the bias-correction algorithm presents a method that can fundamentally eliminate heterogeneity while retaining the benefits of momentum acceleration. Several studies have explored the incorporation of momentum methods into DSGT and their convergence properties \citep{takezawa2023momentum,gao2023distributed}. \cite{huang2024accelerated} introduced a loopless Chebyshev acceleration (LCA) technique to improve the consensus rate and prove its convergence properties under the PL condition case and general non-convex case. Nonetheless, the momentum version of ED/D$^2$, which functions as a bias-correction algorithm, has received limited attention, despite the theoretical results of \cite{Unified_D2_yuan2022} indicating that ED/D$^2$ should exhibit superior convergence properties in sparse networks and greater robustness in the presence of highly heterogeneous data.

Regarding the theoretical properties of the momentum-based algorithms, \cite{liu2020improved} demonstrated that centralized stochastic gradient descent with momentum possesses the same convergence bounds as standard stochastic gradient descent. However, existing theoretical analyses of momentum-based bias-correction algorithms often overlook this property and impose additional constraints on step size or momentum selection. For instance, \cite{takezawa2023momentum,gao2023distributed} require step size $\alpha = \mathcal{O}((1-\lambda)^2)$ and \cite{huang2024accelerated} require the momentum parameter $1 - \beta = \mathcal{O}(1-\lambda)$. In contrast, \cite{Unified_D2_yuan2022} only requires a step size of $\alpha = \mathcal{O}(1-\lambda)$. Therefore, whether momentum-based algorithms can achieve convergence rates comparable to those of the original algorithms remains an important question.

\begin{table}[ht]
\caption{\footnotesize General Non-Convex Convergence Rate. We focus on the condition when $1-\lambda$ is very small and treat the momentum parameter $\beta$ as a constant. Our conclusion regarding EDM eliminates the impact of the network sparsity $\underline{\lambda}$, the minimum non-zero eigenvalue of $\mathbf{W}$, as stated in \cite{Unified_D2_yuan2022}. Additionally, when $\beta = 0$, our conclusion aligns with \cite{yuan2020influence}. Our conclusion demonstrates that the ED/D$^2$ can eliminate the influence of heterogeneity more rapidly than the DSGT, since the heterogeneous term (the \textbf{bold} part) of EDM is scales with $(1-\lambda)^{-2}$, whereas DSGT scales with $(1-\lambda)^{-3}$.}
\centering
\resizebox{\textwidth}{!}{
\begin{tabular}{ccccc}
\toprule
Method & Work & Momentum & Bound of $\alpha$ & Convergence Rate \\ 
\midrule
ED/D$^2$ & \cite{Unified_D2_yuan2022} & - & $\mathcal{O}(1-\lambda)$ & $\mathcal{O}\left(\frac{1}{\alpha T} + \frac{\alpha L\sigma^2}{n} + \frac{\alpha^2L^2\sigma^2}{(1-\lambda)\underline{\lambda}} + \frac{\alpha^4L^4\sigma^2}{(1-\lambda)^3\underline{\lambda}} + \bm{\frac{\alpha^2L^2\zeta_0^2}{(1-\lambda)^2\underline{\lambda}T}}\right)$ \\
Ours & Ours & \checkmark & $\mathcal{O}(1-\lambda)$  & 
$\mathcal{O}\left(\frac{1}{\alpha T} +\frac{\alpha L\sigma^2}{n}+ \frac{\alpha^2L^2\sigma^2}{1-\lambda} + \bm{\frac{\alpha^2L^2\zeta_0^2}{(1-\lambda)^2T}}\right)$ \\
\midrule
DSGT &\cite{Unified_D2_yuan2022} & - & $\mathcal{O}(1-\lambda)$ & $\mathcal{O}\left(\frac{1}{\alpha T} + \frac{\alpha L\sigma^2}{n} + \frac{\alpha^2L^2\sigma^2}{1-\lambda} + \frac{\alpha^4L^4\sigma^2}{(1-\lambda)^4} + \bm{\frac{\alpha^2L^2\zeta_0^2}{(1-\lambda)^3T}}\right)$ \\
DSGT-HB & \cite{gao2023distributed} & \checkmark &  $\mathcal{O}((1-\lambda)^2)$ & 
$\mathcal{O}\left(\frac{1}{\alpha T} + \frac{\alpha L \sigma^2}{n} + \frac{\alpha^2 L^2 \sigma^2}{(1-\lambda)^3} + \bm{\frac{\alpha^2 L^2 \zeta_0^2}{(1-\lambda)^3T}} \right)$\\
\midrule
DmSGD & \cite{yu2019linear}& \checkmark & $\mathcal{O}((1-\lambda)^2)$ & $\mathcal{O}\left(\frac{1}{\alpha T} + \frac{\alpha L \sigma^2}{n} + \frac{\alpha^2L^2\sigma^2}{1-\lambda} + \bm{\frac{\alpha^2L^2\zeta^2}{(1-\lambda)^2}}\right)$\\
DecentLaM & \cite{yuan2021decentlam}& \checkmark & $\mathcal{O}((1-\lambda)^2)$ & $\mathcal{O}\left(\frac{1}{\alpha T} + \frac{\alpha L \sigma^2}{n} + \frac{\alpha^2L^2\sigma^2}{1-\lambda} + \bm{\frac{\alpha^2L^2\zeta^2}{(1-\lambda)^2}}\right)$\\
Quasi-Global & \cite{lin2021quasi}& \checkmark & $\mathcal{O}(1-\lambda)$ & $\mathcal{O}\left(\frac{1}{\alpha T} + \frac{\alpha L \sigma^2}{n} + \frac{\alpha^2L^2\sigma^2}{(1-\lambda)^2} + \bm{\frac{\alpha^2L^2\zeta^2}{(1-\lambda)^2}}\right)$ \\
\bottomrule
\end{tabular}}
\label{table:1}
\end{table}

The contributions of this paper can be summarized as follows.
\begin{enumerate}
    \item We propose a novel decentralized optimization algorithm that incorporates momentum into the ED/D$^2$ framework. This algorithm can accelerate the ED/D$^2$ algorithm in decentralized scenarios. And also ensures that our approach will not be affected by data heterogeneity and the network sparsity and will eventually converge to some neighborhood of the optimal solution, thus being more robust to the data heterogeneity. To the best of our knowledge, this work is the first work to investigate the momentum version of ED/D$^2$ and to provide a theoretical analysis. Table \ref{table:1} shows the theoretical advantages of our algorithm.
    \item This paper introduces a new technique to address the variance introduced by the complicated structure of stochastic gradients. Our method establishes a tight upper bound on the convergence rate, which aligns with the findings of the original ED/D$^2$ algorithm as presented in \cite{Unified_D2_yuan2022}, applicable to both non-convex and PL conditions. This alignment reinforces the conclusions drawn by \cite{liu2020improved}. Furthermore, our proof methodology can be extended to demonstrate the effectiveness of other momentum-based bias-correction algorithms.
\end{enumerate}

This work is arranged as follows. Section 2 addresses the problems to be addressed and the associated challenges. In Section 3, a momentum-based bias-correction algorithm is proposed. Section 4 provides a detailed analysis of the convergence properties of this algorithm. Proofs of relevant theorems are included in the Appendix.
The details and results of our simulation are presented in the electronic companion.

\section{Preliminaries}
\subsection{Notations}
Throughout this paper, we denote the parameter and data of agent $i$ at time $t$ as $\mathbf{x}_i^{(t)}\in \mathbb{R}^{d}$ and $\bm\xi_i^{(t)}\in \mathbb{R}^p$, respectively.  Define $\mathbf{X}^{(t)} = (\mathbf{x}_1^{(t)},\cdots,\mathbf{x}_n^{(t)})^\top \in \mathbb{R}^{n\times d}$ and $\bm\Xi^{(t)} = (\bm\xi_1^{(t)}, \cdots, \bm\xi_n^{(t)})^\top$. For other variables, we use superscripts to indicate parameters or data at time $t$ and subscripts to denote the labels of agents. We put 
$f_i: \mathbb{R}^{d}\to \mathbb{R}$, $\mathbf{f}: \mathbb{R}^{n\times d}\to \mathbb{R}^n$, $\mathbf{f}(\mathbf{X}) = (f_1(\mathbf{x}_1),\cdots,f_n(\mathbf{x}_n))^\top \in \mathbb{R}^n$. $F_i:\mathbb{R}^{d} \times \mathbb{R}^{p} \to \mathbb{R}$, $\mathbf{F}: \mathbb{R}^{n\times d}\to \mathbb{R}^n$, $\mathbf{F}(\mathbf{X},\bm\Xi) = (F_1(\mathbf{x}_1,\bm\xi_1),\cdots,F_n(\mathbf{x}_n,\bm\xi_n))^\top$. The gradient of $\mathbf{F}$ w.r.t. $\mathbf{X}$ is defined as $\nabla \mathbf{F}(\mathbf{X},\bm\Xi) = (\nabla F_1(\mathbf{x}_1,\bm\xi_1),\cdots,\nabla F_n(\mathbf{x}_n,\bm\xi_n))^\top$, and similarly for $\nabla \mathbf{f}(\mathbf{X})$. Let $\mathbf{1}_d = (1,\cdots,1)^\top \in \mathbb{R}^d$ denote the all-ones vector. The mean value of all node parameters is given by $\bar{\mathbf{x}} = \frac1n \mathbf{X}^\top \mathbf{1}_n \in \mathbb{R}^d$. Additionally, we define the matrix $\overline{\mathbf{X}} = \frac1n \mathbf{1}_n\mathbf{1}_n^\top \mathbf{X} \in \mathbb{R}^{n\times d}$. We define the average of loss functions with the local parameter as $\overline{\mathbf{F}}: \mathbb{R}^{n\times d} \times \mathbb{R}^{n\times p} \to \mathbb{R}$, given by $\overline{\mathbf{F}}(\mathbf{X},\bm\Xi) = \frac1n \mathbf{F}(\mathbf{X},\bm\Xi)^\top \mathbf{1}_n$. The mean of the individual loss functions $f_i(\mathbf{x}_i)$ is denoted as $\bar{\mathbf{f}}: \mathbb{R}^{n\times d} \to \mathbb{R}$ with $\bar{\mathbf{f}}(\mathbf{X}) = \frac1n \mathbf{f}(\mathbf{X})^\top \mathbf{1}_n$ The objective function $f: \mathbb{R}^d \to \mathbb{R}$ with $ f(\mathbf{x}) = \frac{1}{n} \sum_{i = 1}^n f_i(\mathbf{x})$. For the momentum, let $\mathbf{m}_i^{(t)}$ be the momentum of agent $i$ at time $t$. Let $\mathbf{M}^{(t)} = (\mathbf{m}_1^{(t)}, \cdots, \mathbf{m}_n^{(t)})^\top$ and $\bar{\mathbf{m}}^{(t)} = \frac{1}{n}(\mathbf{1}_n^\top \mathbf{M}^{(t)})^\top $. Denote the $\ell_2$-norm of $\mathbf{x}\in \mathbb{R}^d$ as $\Vert \mathbf{x}\Vert$, the Frobenius norm of matrix $\mathbf{A}$ as $\Vert\mathbf{A}\Vert_{\mathrm{F}}$ and the spectral norm as
$\Vert\mathbf{A}\Vert_{\mathrm{op}}$.

\subsection{Assumptions}
In the decentralized setting, each agent in the network can only communicate with its neighbors to aggregate information. The network is defined as $\mathcal{G} = (\mathcal{V}, \mathcal{E})$ where the set of nodes is $\mathcal{V} = \{1,\cdots,n\}$, and $\mathcal{E} \subseteq \mathcal{V}\times \mathcal{V}$ represents the connections between neighboring nodes. We denote the neighbors of agent $i$ as $\mathcal{N}_i = \{j: (i,j)\in \mathcal{E}\}$.

\begin{assumption}\label{assump of W}
    Let $\mathbf{W} = (w_{ij})_{1\leq i,j\leq n}$ be the communication matrix corresponding to the network $\mathcal{G}$. The weights $w_{ij}>0$ when $(i,j)\in \mathcal{E}$, and $w_{ij}=0$ otherwise. The matrix $\mathbf{W}$ is designed to satisfy the following conditions:
    \begin{enumerate}
        \item[(1)] For any $i \in \mathcal{V}$, $w_{ii} > 0$;
        \item[(2)] The matrix $\mathbf{W}$ is symmetric and double stochastic, meaning $\mathbf{W}\mathbf{1}_n = \mathbf{1}_n$ and $\mathbf{1}_n^\top \mathbf{W} = \mathbf{1}_n^\top$;
        \item[(3)] The smallest eigenvalue of $\mathbf{W}$ is positive.
    \end{enumerate}
\end{assumption}
\begin{remark}
    Let $\lambda = \Vert\mathbf{W}-\frac1n \mathbf{1} \mathbf{1}^\top\Vert_{\mathrm{op}}$, which is the second largest eigenvalue of $\mathbf{W}$. From (1) and (2) in Assumption \ref{assump of W}, we have $\lambda < 1$ \citep{sayed2014adaptation}. 
   The quantity $(1-\lambda)^{-1}$ is referred to as the spectral gap of the communication matrix $\mathbf{W}$ \citep{nedic2018network,huang2024accelerated}, which is related to the efficiency of the consensus among the agents. The relationship between network structure and spectral gap is discussed in detail in \cite{nedic2018network}. For example, in a ring graph where each agent is arranged in a circle and can only communicate with its two neighbors, the spectral gap of the network will be $O (n^2)$. Generally, as the number of agents increases, the network tends to become sparser, which can lead to a larger spectral gap and subsequently hinder the convergence speed.

    It is worth noting that (1) and (2) in Assumption \ref{assump of W} also imply that the smallest eigenvalue of $\mathbf{W}$ is greater than $-1$ \citep{sayed2014adaptation}. Therefore, for any symmetric doubly stochastic matrix $\mathbf{W}$, Assumption \ref{assump of W} (3) can be satisfied through the transformation $\widetilde{\mathbf{W}} = \frac{\mathbf{W} + \mathbf{I}}{2}$. 
\end{remark}

\begin{assumption}\label{assump of L-smooth}
    {\bf{(Smooth)}} Each local objective function $f_i(\mathbf{x})$ is $L$-smooth:
    \[\left\Vert\nabla f_i(\mathbf{x}) - \nabla f_i(\mathbf{y})\right\Vert\leq L\left\Vert\mathbf{x}-\mathbf{y}\right\Vert, \forall \mathbf{x},\mathbf{y}\in\mathbb{R}^d.\]
    And each $f_i(\mathbf{x})$ is bounded from below, i.e., $f_i(\mathbf{x}) \geq f_i^\star = \inf_{\mathbf{x}} f_i(\mathbf{x}) > -\infty$ for all $\mathbf{x} \in \mathbb{R}^d$.
\end{assumption}
\begin{remark}
    This assumption is widely utilized in distributed stochastic gradient descent algorithms \citep{Unified_D2_yuan2022,huang2024accelerated}, which mainly restricts the objective function to have a certain smoothness, to ensure that each step of gradient descent is relatively gentle.
\end{remark}

\begin{assumption} \label{assump of variance}
    {\bf{(Unbiasedness and finite variance)}} The random data $\bm\xi_i^{(t)}$ is generated independently and identically from $\mathcal{D}_i$. And $\bm\xi_i^{(t)}$ are independent of each other for all $1\leq i\leq n$ and $t\geq 0$. Let $\mathcal{F}^{(t)} = \mathcal{A}(\bm\Xi^{(0)},\cdots,\bm\Xi^{(t-1)})$ denote the sigma-algebra generated by $\bm\Xi^{(0)},\cdots,\bm\Xi^{(t-1)}$. For the $\mathcal{F}^{(t)}$ measurable random vector $\mathbf{x}_i^{(t)}$, the following two formula hold,
    \begin{equation*}
    \begin{aligned}
        &\mathbb{E}[\nabla F_i(\mathbf{x}_i^{(t)},\bm\xi_i^{(t)}) - \nabla f_i(\mathbf{x}_i^{(t)})|\mathcal{F}^{(t)}] = 0,\\
        &\mathbb{E}[\Vert\nabla F_i(\mathbf{x}_i^{(t)},\bm\xi_i^{(t)}) - f_i(\mathbf{x}_i^{(t)})\Vert^2 | \mathcal{F}^{(t)}] \leq \sigma^2.
    \end{aligned}
    \end{equation*}
\end{assumption}
\begin{remark}
    This assumption implies the stochastic gradient at each node is unbiased and possesses a uniform finite variance bound.
\end{remark}

\begin{assumption} \label{assump of PL}
{\bf{(Polyak-\L ojasiewicz condition)}} There exists a constant $\mu > 0$ such that for all $\mathbf{x}\in \mathbb{R}^d$, the objective function $f(\mathbf{x}) = \frac{1}{n}\sum_{i = 1}^n f_i(\mathbf{x})$ satisfies
\[2\mu(f(\mathbf{x}) - f^\star) \leq \left\Vert\nabla f(\mathbf{x})\right\Vert^2.\]
where $f^\star$ denotes the optimal value of $f$.
\end{assumption}
\begin{remark}
    This assumption is widely used in recent theoretical analyses of decentralized algorithms, e.g. \cite{Unified_D2_yuan2022,huang2024accelerated}. And the PL-condition is weaker than the $\mu$-strongly convex assumption, which requires $f(\mathbf{y}) - f(\mathbf{x}) -\langle\nabla f(\mathbf{x}),\mathbf{y}-\mathbf{x}\rangle \geq \frac{\mu}{2}\left\Vert\mathbf{x}-\mathbf{y}\right\Vert^2$. Many studies on decentralized SGD with momentum acceleration \citep{lin2021quasi,yuan2021decentlam} assume that each agent's objective function is strongly convex $f_i(\mathbf{x})$.
\end{remark}

\section{Proposed Method}
\subsection{Drawbacks of momentum-based DSGD algorithm}

Decentralized momentum SGD (DmSGD) \citep{yu2019linear,gao2020periodic} was proposed to accelerate the convergence of distributed learning. In DmSGD, each agent $i$ iteratively updates its momentum and communicates parameters as follows:
\begin{align}
    \mathbf{m}_i^{(t+1)} & = \beta \mathbf{m}_i^{(t)} + (1-\beta) \nabla F_i(\mathbf{x}_i^{(t)}, \bm\xi_{i}^{(t)}),\label{eq: momentum updater}\\
    \mathbf{x}_i^{(t+1)} & = \sum_{j=1}^n w_{ij} \left(\mathbf{x}_i^{(t)}-\alpha \mathbf{m}_i^{(t+1)}\right).\label{dmsgd} 
\end{align}
However, Proposition 2 in \cite{yuan2021decentlam} demonstrated that DmSGD suffers from an inconsistency bias for strongly convex loss under full-batch gradients (i.e. when $\sigma^2 = 0$),
\[\lim_{t\to \infty} \sum_{i=1}^n\left\Vert\mathbf{x}_i^{(t)}-\mathbf{x}^*\right\Vert^2 = \mathcal{O}\left(\frac{\alpha^2 \zeta^2}{(1-\beta)^2(1-\lambda)^2}\right),\]
where $\mathbf{x}^\star$ denotes the optimum point, and $\zeta^2 = (1/n) \sum_{i=1}^n\left\Vert\nabla f_i(\mathbf{x}^*) - \nabla f(\mathbf{x}^*)\right\Vert^2$ is defined as the measure of the data heterogeneity \citep{hsieh2020non,koloskova2020unified}. Additionally, through the proof of theorem 3 in \cite{yu2019linear}, it can be shown that for the non-convex objective function $f_i(\mathbf{x})$, DmSGD will also converge to a neighborhood of local optimal, with the radius being influenced by the variance of the stochastic gradient $\sigma^2$ and the data heterogeneity $\zeta^2$. When the data exhibit greater heterogeneity, DmSGD tends to oscillate over a larger area surrounding the optimum, resulting in larger corresponding errors. Therefore, addressing the challenge of leveraging momentum to accelerate the decentralized algorithm while simultaneously reducing heterogeneity is an urgent problem that needs to be resolved.

\subsection{From Exact-Diffusion to Proposed Method}
The challenge of heterogeneity is inherent in distributed SGD. To ensure the stability of the convergence, we need to guarantee that the parameter at the optimal value point remains in the optimal position after one step of iteration. However, the heterogeneity of data commonly leads to the condition that $\nabla f_i(\mathbf{X}^\star) \neq 0$ (where $\mathbf{X}^\star = \mathbf{1} {\mathbf{x}^\star}^\top$ and $\mathbf{x}^\star$ denotes the optimal point). Consequently, the algorithm will move away from the optimal value after one step gradient update, resulting in $\mathbf{X}^{(t+1)} = \mathbf{X}^\star - \alpha \nabla \mathbf{f}(\mathbf{X}^\star) \neq \mathbf{X}^\star$. Consequently, it is essential to employ specific methods to correct the biases introduced by the inconsistency in the gradient. A class of bias-correction distributed SGD algorithms has been developed to address heterogeneity, summarized as the SUDA algorithm in \cite{Unified_D2_yuan2022}. In this paper, we focus on a specific class of algorithms based on the ED/D$^2$ framework, which can be summarized in the following three steps: first, perform local gradient descent represented by $\bm\psi_i^{(t+1)} = \mathbf{x}_i^{(t)} - \alpha \nabla{F}_i(\mathbf{x}_i^{(t)},\bm\xi_i^{(t)})$, where we use $\bm\psi_i^{(t)}$ to denote the temporary variable. Second, correct the bias caused by the data heterogeneity as expressed by $\bm\phi_i^{(t+1)} = \bm\psi_i^{(t+1)} + \mathbf{x}_i^{(t)} - \bm\psi_i^{(t)}$. Finally, combine the corrected parameter $\bm\phi_i^{(t+1)}$ to update the local parameter with $\mathbf{x}_i^{(t+1)} = \sum_{j\in\mathcal{N}_i} w_{ij} \bm\phi_j^{(t+1)}$, where $\mathcal{N}_i = \{j: (i,j)\in \mathcal{E}\}$ represents the set of neighbors for agent $i$.

The main idea of ED/D$^2$ is to utilize local historical information to correct the bias introduced by the data heterogeneity during the local gradient update step. Thus, ED/D$^2$ will ultimately converge to a neighborhood of the optimum, influenced solely by the randomness inherent in the stochastic gradient. Generally, the iteration of ED/D$^2$ can be expressed as follows:
\begin{equation*}
    \begin{aligned}
        \mathbf{X}^{(t+2)} = & \mathbf{W}\left(2\mathbf{X}^{(t+1)} - \mathbf{X}^{(t)} - \alpha \nabla\mathbf{F}(\mathbf{X}^{(t+1)},\bm\Xi^{(t+1)}) + \alpha\nabla\mathbf{F}(\mathbf{X}^{(t)},\bm\Xi^{(t)})\right).
    \end{aligned}
\end{equation*}

To incorporate the idea of momentum acceleration into ED/D$^2$, we replace $\nabla F_i(\mathbf{x}_i^{(t)},\bm\xi_i^{(t)})$ with a weighted sum over the historical gradient, specifically $\mathbf{m}_i^{(t)}$ from DmSGD as defined in \eqref{eq: momentum updater}. This leads to the development of the \textbf{E}xact-\textbf{D}iffusion with \textbf{M}omentum (\textbf{EDM}) algorithm, which can be formally described in Algorithm \ref{alg:EDM}. The iteration formula can be written as:
\begin{equation}\label{eq: def of x}
    \mathbf{X}^{(t+2)} = \mathbf{W}\left(2\mathbf{X}^{(t+1)} - \mathbf{X}^{(t)} - \alpha\mathbf{M}^{(t+1)} + \alpha\mathbf{M}^{(t)}\right).
\end{equation}

\begin{algorithm}[ht]
\caption{Exact Diffusion with Momentum (EDM)}
\vskip6pt
\label{alg:EDM}
\begin{algorithmic}
        \State \textbf{Initialization:} Generate the initial value $\mathbf{x}^{(0)}$ from data or randomly and let $\mathbf{x}_i^{(-1)} = \mathbf{x}_i^{(0)} = \mathbf{x}^{(0)}$ at each node $i$. Initialize the parameters $\alpha$ and the double stochastic matrix $\mathbf{W}$.
        \For{$t=0\cdots T-1$}
            \State {Estimate the local stochastic gradient based on $\bm\xi_i^{(t)}$: $\mathbf{g}_i^{(t)} = \nabla F_i(\mathbf{x}_i^{(t)},\bm\xi_i^{(t)})$};
            \State {Momentum update: $\mathbf{v}_i^{(t)} = \beta \mathbf{m}_i^{(t-1)} + (1-\beta)\mathbf{g}_i^{(t)}  $};
            \State {Adapt: $\bm\psi_i^{(t+1)} = \mathbf{x}_i^{(t)} - \alpha \mathbf{m}^{(t)}$};
            \State {Correct: $\bm\phi_i^{(t+1)} = \bm\psi_i^{(t+1)} + \mathbf{x}_i^{(t)} - \bm\psi_i^{(t)}$};
            \State {Combine: 
            $\mathbf{x}_i^{(t+1)} = \sum_{j\in\mathcal{N}_i} w_{ij} \bm\phi_j^{(t+1)} $};
        \EndFor
        \State \textbf{Output:} $\mathbf{x}_i^{(T)}$ for all agent $i \in \mathcal{V}$.
    \end{algorithmic} 
\end{algorithm}

Assume that $\mathbf{X}^{(t)}$ is obtained by Algorithm \ref{alg:EDM}, and $\bar{\mathbf{x}}^{(t)} = \frac{1}{n}\mathbf{X}^{(t)}\mathbf{1}_n$, we have $\bar{\mathbf{x}}^{(t+1)} = \bar{\mathbf{x}}^{(t)} - \alpha \bar{\mathbf{m}}^{(t)}$. This implies that although the parameters differ across agents, the EDM algorithm exhibits the same update pattern as standard momentum SGD in the average sense.

\section{Convergence Analysis}

\subsection{Auxiliary Variables}
To measure the convergence rate, we primarily focus on $f(\bar{\mathbf{x}}^{(t)})$ and its gradient $\nabla f(\bar{\mathbf{x}}^{(t)})$. We begin by defining an auxiliary sequence for $\bar{\mathbf{x}}^{(t)}$, which is commonly used in the analysis of momentum-based SGD methods. Let $\mathbf{z}^{(0)} = \bar{\mathbf{x}}^{(0)}$ and define
\begin{equation}\label{eq: def of z}
    \mathbf{z}^{(t)} = \frac{1}{1-\beta} \bar{\mathbf{x}}^{(t)} - \frac{\beta}{1-\beta} \bar{\mathbf{x}}^{(t-1)}, \quad t\geq 1.
\end{equation}

The following two lemmas establish the connection between $\mathbf{z}^{(t)}$ and $\mathbf{X}^{(t)}$ under the non-convex cases and PL conditions, respectively.
\begin{lemma}\label{lem: iter for f nonconvex}
Suppose that Assumptions \ref{assump of W}-\ref{assump of variance} hold. We have the following inequality holds for $t\geq 0$,
\begin{equation}\label{eq: iter for nabla barf}
\begin{aligned}
    \mathbb{E}[f(\mathbf{z}^{(t+1)})] \leq & \mathbb{E}[f(\mathbf{z}^{(t)})] + \frac{\alpha^2L\beta}{2(1-\beta)}\mathbb{E}\Vert\bar{\mathbf{m}}^{(t-1)}\Vert^2 + \frac{\alpha L^2}{2n}\mathbb{E}\left\Vert\mathbf{X}^{(t)} - \overline{\mathbf{X}}^{(t)} \right\Vert_{\mathrm{F}}^2 + \frac{\alpha^2L\sigma^2}{2n} \\
    & - \frac{\alpha}2\left(1-\frac{\beta\alpha L}{1-\beta} - \alpha L\right)\mathbb{E}\left\Vert\nabla \bar{\mathbf{f}}(\mathbf{X}^{(t)})\right\Vert^2 - \frac{\alpha}2 \mathbb{E}\left\Vert\nabla \bar{\mathbf{f}}(\overline{\mathbf{X}}^{(t)})\right\Vert^2 .
\end{aligned}
\end{equation}
\end{lemma}

\begin{lemma}\label{lem: iter for f PL}
Let $\tilde{f}(\mathbf{x}) = \sum_{i = 1}^n f_i(\mathbf{x}) - f^\star$. Under Assumptions \ref{assump of W}-\ref{assump of PL}, we have
\begin{equation}\label{eq: iter for nabla barf PL}
\begin{aligned}
    \mathbb{E}[\tilde{f}(\mathbf{z}^{(t+1)})] \leq & (1-\alpha\mu)\mathbb{E}[\tilde{f}(\mathbf{z}^{(t)})]  - \frac{\alpha(1-\alpha L)}{2}\mathbb{E}\left\Vert\nabla \bar{\mathbf{f}}(\mathbf{X}^{(t)})\right\Vert^2 \\
    & + \frac{\alpha L^2}{2n}\mathbb{E}\left\Vert\mathbf{X}^{(t)} - \overline{\mathbf{X}}^{(t)}\right\Vert_{\mathrm{F}}^2 + \frac{\alpha^3 L^2\beta^2}{2(1-\beta)^2}\mathbb{E}\left\Vert\bar{\mathbf{m}}^{(t-1)}\right\Vert^2 + \frac{\alpha^2L\sigma^2}{2n}.
\end{aligned}
\end{equation} 
\end{lemma}

\subsection{Pseudo Deviation Variance Decomposition}
According to Lemma \ref{lem: iter for f nonconvex} and \ref{lem: iter for f PL}, our next step is to conduct an in-depth investigation of $\mathbb{E}\Vert\mathbf{X}^{(t)} - \overline{\mathbf{X}}^{(t)}\Vert_{\mathrm{F}}^2$. To simplify the notation, we introduce $\mathbf{P}_\mathbf{I} = \mathbf{I} - \frac1n  \mathbf{1}_n\mathbf{1}_n^\top$, thus $\Vert\mathbf{X}^{(t)} - \overline{\mathbf{X}}^{(t)}\Vert_{\mathrm{F}}^2 = \Vert\mathbf{P}_\mathbf{I}\mathbf{X}^{(t)}\Vert_{\mathrm{F}}^2$. 

The momentum updating structure complicates the exploitation of the independence of $\bm\Xi^{(t)}$ in the stochastic gradient. To address this challenge, we introduce the following variables: let $\widetilde{\mathbf{M}}^{(-1)} = \mathbf{0}$ and $\widetilde{\mathbf{M}}^{(t)} = (1-\beta)\sum_{j = 0}^{k}\beta^{t-j}\nabla \mathbf{f}(\mathbf{X}^{(j)})$, $\widetilde{\mathbf{X}}^{(-1)} = \widetilde{\mathbf{X}}^{(0)} = \mathbf{X}^{(0)}$ and
\begin{equation}\label{eq: def of tildex}
    \widetilde{\mathbf{X}}^{(t+1)} = \mathbf{W}\left(2\widetilde{\mathbf{X}}^{(t)} - \widetilde{\mathbf{X}}^{(t-1)} - \alpha\widetilde{\mathbf{M}}^{(t)} + \alpha\widetilde{\mathbf{M}}^{(t-1)}\right).
\end{equation}
By applying Cauchy’s inequality, we obtain:
\begin{equation}\label{eq: basic deviation variance decomposition}
\begin{aligned}
    \left\Vert\mathbf{P}_{\mathbf{I}}\mathbf{X}^{(t)}\right\Vert_{\mathrm{F}}^2 \leq 2\left\Vert\mathbf{P}_{\mathbf{I}}\widetilde{\mathbf{X}}^{(t)}\right\Vert_{\mathrm{F}}^2 + 2\left\Vert\mathbf{P}_{\mathbf{I}}(\mathbf{X}^{(t)} - \widetilde{\mathbf{X}}^{(t)})\right\Vert_{\mathrm{F}}^2.
\end{aligned}
\end{equation}

Next, we present the upper bound of the noise term.
\begin{lemma}\label{lem: variance bound}
    Suppose that Assumptions \ref{assump of W} and \ref{assump of variance} hold. Let $\mathbf{X}^{(t)}$ is the sequence of parameters obtained from Algorithm \ref{alg:EDM} and $\widetilde{\mathbf{X}}^{(t)}$ be defined as above. Then, we have:
    \begin{equation}\label{eq: bound of PI(x-tildex)}
        \mathbb{E}\left\Vert\mathbf{P}_{\mathbf{I}}(\mathbf{X}^{(t)} - \widetilde{\mathbf{X}}^{(t)})\right\Vert_{\mathrm{F}}^2 \leq \frac{13\alpha^2\lambda^2n\sigma^2}{1-\lambda}.
    \end{equation}
\end{lemma}

This lemma provides an upper bound for $\mathbb{E} \Vert\mathbf{P}_{\mathbf{I}}(\mathbf{X}^{(t)} - \widetilde{\mathbf{X}}^{(t)})\Vert_{\mathrm{F}}^2$, which is independent of $\beta$. Thus, we should only focus on the pseudo-deterministic part $\Vert\mathbf{P}_{\mathbf{I}}\widetilde{\mathbf{X}}^{(t)}\Vert_{\mathrm{F}}^2$. We call it pseudo-deterministic because it eliminates randomness in the gradients while still retaining randomness in $\mathbf{X}^{(t)}$.

\subsection{Transformations and Consensus Inequalities}
Building on the ideas presented in \cite{Unified_D2_yuan2022}, we can observe that $\widetilde{\mathbf{X}}^{(t)}$ exhibits a recursive structure similar to that of the SUDA framework described in their work. Specifically, we define $\widetilde{\mathbf{X}}^{(0)} = \mathbf{X}^{(0)}$, $\widetilde{\mathbf{M}}^{(0)} = (1-\beta)\nabla \mathbf{f}(\mathbf{X}^{(0)})$, $\widetilde{\mathbf{M}}^{(-1)} = \mathbf{0}$, $\widetilde{\mathbf{Y}}^{(0)} = \mathbf{0}$. For all $t\geq 0$, we have:
\begin{align}
    \widetilde{\mathbf{X}}^{(t+1)} &= \mathbf{W}(\widetilde{\mathbf{X}}^{(t)} - \alpha \widetilde{\mathbf{M}}^{(t)}) - (\mathbf{I} - \mathbf{W})^{1/2}\widetilde{\mathbf{Y}}^{(t)},\label{eq: iter for tildexyM(x)}\\
    \widetilde{\mathbf{Y}}^{(t+1)} &= \widetilde{\mathbf{Y}}^{(t)} + (\mathbf{I} - \mathbf{W})^{1/2}\widetilde{\mathbf{X}}^{(t+1)},\label{eq: iter for tildexyM(y)}\\
    \widetilde{\mathbf{M}}^{(t+1)} & = \beta \widetilde{\mathbf{M}}^{(t)}+(1-\beta) \nabla \mathbf{f}(\widetilde{\mathbf{X}}^{(t+1)}).\label{eq: iter for tildexyM(M)}
\end{align}

The following lemma presents a transformation of $\widetilde{\mathbf{X}}^{(t)}$, which is used to derive the upper bound of $\Vert\mathbf{P}_\mathbf{I} \widetilde{\mathbf{X}}^{(t)}\Vert^2$, and consequently for $\Vert\mathbf{P}_\mathbf{I} \mathbf{X}^{(t)}\Vert^2$. This lemma establishes an upper bound on the data consensus deviation that occurs during the iteration, which is central to our analysis of the convergence properties of algorithms. 
\begin{lemma}\label{Lem: transformation for x}
    Let $\widetilde{\mathbf{N}}^{(t)}$ be any series that satisfies $\widetilde{\mathbf{N}}^{(0)} = \nabla \mathbf{f}(\overline{\mathbf{X}}^{(0)})$ and $\widetilde{\mathbf{R}}^{(t)} = \widetilde{\mathbf{M}}^{(t)} - \widetilde{\mathbf{N}}^{(t)}$. Under the conditions specified in assumptions \ref{assump of W}-\ref{assump of variance}, there exists a series of matrices $\{\mathbf{E}^{(t)}\}_{t\geq 1}$, satisfies the following two inequality
    \begin{equation}\label{eq: iter for ek lemma version}
    \begin{aligned}
        \left\Vert\mathbf{E}^{(t+1)}\right\Vert_{\mathrm{F}}^2& \leq \sqrt{\lambda}\left\Vert\mathbf{E}^{(t)}\right\Vert_{\mathrm{F}}^2 + \frac{2\alpha^2 \lambda}{1-\sqrt{\lambda}} \left\Vert\mathbf{P}_{\mathbf{I}} \widetilde{\mathbf{R}}^{(t)}\right\Vert_{\mathrm{F}}^2 + \frac{2\alpha^2 \lambda}{(1-\sqrt{\lambda})(1-\lambda)}\left\Vert\mathbf{P}_{\mathbf{I}}(\widetilde{\mathbf{N}}^{(t+1)} - \widetilde{\mathbf{N}}^{(t)})\right\Vert_{\mathrm{F}}^2,
    \end{aligned}
    \end{equation}
    \begin{equation}\label{eq: ineq between x and e}
        \mathbb{E}\left\Vert\mathbf{P}_{\mathbf{I}}\mathbf{X}^{(t)}\right\Vert_{\mathrm{F}}^2 \leq 8\mathbb{E}\left\Vert\mathbf{E}^{(t)}\right\Vert_{\mathrm{F}}^2 + \frac{26\alpha^2\lambda^2n\sigma^2}{1-\lambda}.
    \end{equation}
\end{lemma}

\begin{remark}
    In this Lemma, we do not specify the exact form of $\widetilde{\mathbf{N}}^{(t)}$. Because taking different values for $\widetilde{\mathbf{N}}^{(t)}$ in different proofs can simplify our proof, even though all forms are intended to measure the difference between the current updates and $\nabla \mathbf{f}(\overline{\mathbf{X}}^{(t)})$. In the subsequent proofs, we will adopt $\widetilde{\mathbf{N}}^{(t)} = (1-\beta)\sum_{j = 0}^{t}\beta^{t-j}\nabla \mathbf{f}(\overline{\mathbf{X}}^{(j)}) + \beta^{t+1}\nabla \mathbf{f}(\overline{\mathbf{X}}^{(0)})$ for the non-convex case and $\widetilde{\mathbf{N}}^{(t)} = \nabla \mathbf{f}(\overline{\mathbf{X}}^{(t)})$ for the PL condition.
\end{remark}

\subsection{Non-Convex}
The following Theorem presents the convergence properties under non-convex conditions.
\begin{theorem}\label{thm: non-convex}
Under Assumptions \ref{assump of W}-\ref{assump of variance}, if the step size $\alpha$ satisfies $\alpha \leq \min\left\{\frac{1-\sqrt{\lambda}}{4L},\frac{1-\beta}{4L}\right\}$, we have
\begin{equation*}
    \begin{aligned}
        & \frac{1}{T}\sum_{t = 0}^{T-1} \bigg(\frac{1}4 \mathbb{E}\left\Vert\nabla \bar{\mathbf{f}}(\mathbf{X}^t))\right\Vert^2 + \mathbb{E}\left\Vert\nabla \bar{\mathbf{f}}(\overline{\mathbf{X}}^t))\right\Vert^2\bigg)\\
        \leq & \frac{2}{\alpha T}\left(f(\mathbf{x}^{(0)}) - f^\star\right)+ \frac{2\alpha L\sigma^2}{n} + \frac{52\alpha^2L^2\lambda^2 \sigma^2}{1-\lambda} + \frac{8C_0\alpha^2L^2\zeta_0^2}{(1-\sqrt{\lambda})^2T}.
    \end{aligned}
\end{equation*}
where $\zeta_0^2 = \frac{1}{n}\left\Vert\mathbf{W}\left(\mathbf{I} - \frac{1}{n} \mathbf{1}\mathbf{1}^\top\right) \nabla \mathbf{f}(\mathbf{X}^{(0)})\right\Vert_{\mathrm{F}}$, $f^\star = \min_x f(x)$ and $C_0 = \frac{24(1-\beta)^2}{1+\sqrt{\lambda}} + 12\beta^2\lambda(1-\sqrt{\lambda}) + \frac{2\beta^3\lambda}{1-\beta}$.
\end{theorem}

In practical optimization problems, we typically select $\beta \in [0.8,0.99)$, allowing us to treat $\beta$ as a fixed value. When $\beta = 0$, the algorithm simplifies to the ED/D$^2$ method. Theorem \ref{thm: non-convex} aligns with the latest conclusions of the ED/D$^2$ framework discussed in \cite{Unified_D2_yuan2022} with the corresponding bounds illustrated in Table \ref{table:1}. In contrast to their findings, we obtained a bound that is unaffected by the minimum eigenvalue of the communication matrix through a refined scaling of the initial values. Consequently, we achieve a tighter upper bound when $\beta = 0$. This results in the momentum-based algorithm exhibiting a convergence rate that is at least comparable to the ED/D$^2$ algorithm. This feature shows the strength of our algorithm.

\begin{remark}
Unlike Quasi-Global and DecentLaM, which are also the momentum-based variants of distributed stochastic gradient descent, EDM can ultimately eliminate the effects of data heterogeneity. Compared with another kind of bias-correction algorithm (DSGT and DSGT-HB), EDM can eliminate data heterogeneity at a rate of $O(\alpha^2(1-\lambda)^{-2}T^{-1})$, while DSGT at a rate of $O(\alpha^2(1-\lambda)^{-3}T^{-1})$. This distinction will be demonstrated in our subsequent real-data analysis.
\end{remark}

\subsection{PL-Condition}
\begin{theorem}\label{thm: PL}
Suppose Assumptions 1-4 hold, if the step size $\alpha$ in the EDM algorithm satisfies $\alpha \leq \min\{\frac{1-\sqrt{\lambda}}{10},\frac{1-\beta}{5}\}$, we have
\begin{equation*}
    \begin{aligned}
        & \mathbb{E} f(\bar{\mathbf{x}}^{(t)}) \leq \left(9\tilde{f}(\mathbf{x}^{(0)}) + \frac{D_1\alpha^3L^2 \zeta_0^2}{(1-\sqrt{\lambda})^2}\right)\rho_1^{t} + \frac{D_2\alpha^2L\zeta_0^2}{1-\lambda}\rho_2^t +\frac{6\alpha L\sigma^2}{n\mu} + \frac{169\alpha^2L^2\lambda^2\sigma^2}{\mu(1-\lambda)}.
    \end{aligned}
\end{equation*}

where $\tilde{f}(\mathbf{x}) = f(\mathbf{x}) - f^\star$, $\rho_1 = 1-\alpha\mu$, $\rho_2 = 1-\min\left\{\frac{1-\sqrt{\lambda}}{5},\frac{2(1-\beta)}{5}\right\}$, and 
\[D_1 = \frac{50\lambda\beta^2}{3(1-\beta)} + \frac{320(1-\beta)^2}{3(1+\sqrt{\lambda})} + \frac{160}{3}\lambda\beta^2(1-\sqrt{\lambda}),\]
\[D_2 = \frac{100}{3}\left(\frac{5\lambda\beta^2}{3(1-\beta)} + 4(1-\beta)^2 + 2\lambda \beta^2 (1-\lambda)\right).\]
\end{theorem}

Similarly to the analysis in the non-convex case, in the PL condition case, when $\beta = 0$, our algorithm can degenerate to ED/D$^2$. Moreover, our algorithm inherits the property of ED/D$^2$, enabling it to linearly eliminate data heterogeneity.
\begin{remark}
    We compare EDM to other momentum-based distributed algorithms under the PL condition. Both the convergence bounds of Quasi-Global and DecentLaM fail to eliminate the effect of heterogeneity. Currently, no work proves the complete convergence properties of the DSGT algorithm with momentum under the Polyak-Lojasiewicz (PL) condition, or even under the strong convexity condition. While \cite{huang2024accelerated} demonstrates convergence properties when $\beta$ is sufficiently large, this result does not extend to the original algorithm (i.e. when $\beta = 0$). In contrast, our findings impose no constraints on the choice of $\beta$ and encompass the conclusions of ED/D$^2$. Furthermore, our approach can be generalized to establish convergence properties for momentum-based DSGT algorithms.
    
    On the other hand, \cite{huang2024accelerated} employs the technique of Loopless Chebyshev Acceleration (LCA) to reduce the communication gap in DSGT from $(1-\lambda)^{-1}$ to $(1-\lambda)^{-1/2}$, resulting in a better upper bound on convergence. However, this technique is not exclusive to DSGT; other distributed algorithms can also benefit from LCA to reduce the communication gap. But this aspect is not the focus of our article.
\end{remark}

\section{Conclusions}\label{sec:Conclusion}
This paper focuses on the acceleration algorithm of decentralized stochastic gradient descent in sparse communication networks and under conditions of data heterogeneity. By incorporating the idea of momentum acceleration into the bias-correction decentralized stochastic gradient algorithm, we propose a new algorithm that achieves consensus more rapidly in the presence of heterogeneous data. We theoretically demonstrate that, under non-convex and PL conditions, the proposed algorithm exhibits at least comparable convergence properties to the original algorithm, effectively mitigating the impact of data heterogeneity. 

However, the specific mechanisms of momentum acceleration remain unclear. Therefore, further investigation is necessary to establish a robust framework for distributed momentum acceleration and, more broadly, for centralized momentum SGD.

\vskip 0.2in

\bibliographystyle{chicago}

\newpage
\bibliography{references}

\begin{thebibliography}{}

\bibitem[\protect\citeauthoryear{Alghunaim and Yuan}{Alghunaim and Yuan}{2022}]{Unified_D2_yuan2022}
Alghunaim, S.~A. and K.~Yuan (2022).
\newblock A unified and refined convergence analysis for non-convex decentralized learning.
\newblock {\em IEEE Transactions on Signal Processing\/}~{\em 70}, 3264--3279.

\bibitem[\protect\citeauthoryear{Alghunaim and Yuan}{Alghunaim and Yuan}{2024}]{alghunaim2024enhanced}
Alghunaim, S.~A. and K.~Yuan (2024).
\newblock An enhanced gradient-tracking bound for distributed online stochastic convex optimization.
\newblock {\em Signal Processing\/}~{\em 217}, 109345.

\bibitem[\protect\citeauthoryear{Ali, Vecchio, Pincheira, Dolui, Antonelli, and Rehmani}{Ali et~al.}{2018}]{ali2018applications}
Ali, M.~S., M.~Vecchio, M.~Pincheira, K.~Dolui, F.~Antonelli, and M.~H. Rehmani (2018).
\newblock Applications of blockchains in the internet of things: A comprehensive survey.
\newblock {\em IEEE Communications Surveys \& Tutorials\/}~{\em 21\/}(2), 1676--1717.

\bibitem[\protect\citeauthoryear{Beltr{\'a}n, P{\'e}rez, S{\'a}nchez, Bernal, Bovet, P{\'e}rez, P{\'e}rez, and Celdr{\'a}n}{Beltr{\'a}n et~al.}{2023}]{beltran2023decentralized}
Beltr{\'a}n, E. T.~M., M.~Q. P{\'e}rez, P.~M.~S. S{\'a}nchez, S.~L. Bernal, G.~Bovet, M.~G. P{\'e}rez, G.~M. P{\'e}rez, and A.~H. Celdr{\'a}n (2023).
\newblock Decentralized federated learning: Fundamentals, state of the art, frameworks, trends, and challenges.
\newblock {\em IEEE Communications Surveys \& Tutorials\/}.

\bibitem[\protect\citeauthoryear{Boyd, Parikh, Chu, Peleato, Eckstein, et~al.}{Boyd et~al.}{2011}]{boyd2011distributed}
Boyd, S., N.~Parikh, E.~Chu, B.~Peleato, J.~Eckstein, et~al. (2011).
\newblock Distributed optimization and statistical learning via the alternating direction method of multipliers.
\newblock {\em Foundations and Trends{\textregistered} in Machine learning\/}~{\em 3\/}(1), 1--122.

\bibitem[\protect\citeauthoryear{Gao and Huang}{Gao and Huang}{2020}]{gao2020periodic}
Gao, H. and H.~Huang (2020).
\newblock Periodic stochastic gradient descent with momentum for decentralized training.
\newblock {\em arXiv preprint arXiv:2008.10435\/}.

\bibitem[\protect\citeauthoryear{Gao, Liu, Dai, Huang, and Gu}{Gao et~al.}{2023}]{gao2023distributed}
Gao, J., X.-W. Liu, Y.-H. Dai, Y.~Huang, and J.~Gu (2023).
\newblock Distributed stochastic gradient tracking methods with momentum acceleration for non-convex optimization.
\newblock {\em Computational Optimization and Applications\/}~{\em 84\/}(2), 531--572.

\bibitem[\protect\citeauthoryear{Hsieh, Phanishayee, Mutlu, and Gibbons}{Hsieh et~al.}{2020}]{hsieh2020non}
Hsieh, K., A.~Phanishayee, O.~Mutlu, and P.~Gibbons (2020).
\newblock The non-iid data quagmire of decentralized machine learning.
\newblock In {\em International Conference on Machine Learning}, pp.\  4387--4398. PMLR.

\bibitem[\protect\citeauthoryear{Huang, Pu, and Nedi{\'c}}{Huang et~al.}{2024}]{huang2024accelerated}
Huang, K., S.~Pu, and A.~Nedi{\'c} (2024).
\newblock An accelerated distributed stochastic gradient method with momentum.
\newblock {\em arXiv preprint arXiv:2402.09714\/}.

\bibitem[\protect\citeauthoryear{Jiang, Balu, Hegde, and Sarkar}{Jiang et~al.}{2017}]{jiang2017collaborative}
Jiang, Z., A.~Balu, C.~Hegde, and S.~Sarkar (2017).
\newblock Collaborative deep learning in fixed topology networks.
\newblock {\em Advances in Neural Information Processing Systems\/}~{\em 30}.

\bibitem[\protect\citeauthoryear{Koloskova, Loizou, Boreiri, Jaggi, and Stich}{Koloskova et~al.}{2020}]{koloskova2020unified}
Koloskova, A., N.~Loizou, S.~Boreiri, M.~Jaggi, and S.~Stich (2020).
\newblock A unified theory of decentralized sgd with changing topology and local updates.
\newblock In {\em International Conference on Machine Learning}, pp.\  5381--5393. PMLR.

\bibitem[\protect\citeauthoryear{Lian, Zhang, Zhang, Hsieh, Zhang, and Liu}{Lian et~al.}{2017}]{lian2017can}
Lian, X., C.~Zhang, H.~Zhang, C.-J. Hsieh, W.~Zhang, and J.~Liu (2017).
\newblock Can decentralized algorithms outperform centralized algorithms? a case study for decentralized parallel stochastic gradient descent.
\newblock {\em Advances in Neural Information Processing Systems\/}~{\em 30}.

\bibitem[\protect\citeauthoryear{Lin, Karimireddy, Stich, and Jaggi}{Lin et~al.}{2021}]{lin2021quasi}
Lin, T., S.~P. Karimireddy, S.~Stich, and M.~Jaggi (2021).
\newblock Quasi-global momentum: Accelerating decentralized deep learning on heterogeneous data.
\newblock In {\em International Conference on Machine Learning}, pp.\  6654--6665. PMLR.

\bibitem[\protect\citeauthoryear{Liu, Gao, and Yin}{Liu et~al.}{2020}]{liu2020improved}
Liu, Y., Y.~Gao, and W.~Yin (2020).
\newblock An improved analysis of stochastic gradient descent with momentum.
\newblock {\em Advances in Neural Information Processing Systems\/}~{\em 33}, 18261--18271.

\bibitem[\protect\citeauthoryear{Nedi{\'c}, Olshevsky, and Rabbat}{Nedi{\'c} et~al.}{2018}]{nedic2018network}
Nedi{\'c}, A., A.~Olshevsky, and M.~G. Rabbat (2018).
\newblock Network topology and communication-computation tradeoffs in decentralized optimization.
\newblock {\em Proceedings of the IEEE\/}~{\em 106\/}(5), 953--976.

\bibitem[\protect\citeauthoryear{Nedic and Ozdaglar}{Nedic and Ozdaglar}{2009}]{nedic2009distributed}
Nedic, A. and A.~Ozdaglar (2009).
\newblock Distributed subgradient methods for multi-agent optimization.
\newblock {\em IEEE Transactions on Automatic Control\/}~{\em 54\/}(1), 48--61.

\bibitem[\protect\citeauthoryear{Nesterov}{Nesterov}{2013}]{nesterov2013introductory}
Nesterov, Y. (2013).
\newblock {\em Introductory lectures on convex optimization: A basic course}, Volume~87.
\newblock Springer Science \& Business Media.

\bibitem[\protect\citeauthoryear{Polyak}{Polyak}{1964}]{polyak1964some}
Polyak, B.~T. (1964).
\newblock Some methods of speeding up the convergence of iteration methods.
\newblock {\em Ussr Computational Mathematics and Mathematical Physics\/}~{\em 4\/}(5), 1--17.

\bibitem[\protect\citeauthoryear{Polyak}{Polyak}{1987}]{polyak1987introduction}
Polyak, B.~T. (1987).
\newblock {\em Introduction to optimization}.
\newblock New York, Optimization Software.

\bibitem[\protect\citeauthoryear{Pu and Nedi{\'c}}{Pu and Nedi{\'c}}{2021}]{pu2021distributed}
Pu, S. and A.~Nedi{\'c} (2021).
\newblock Distributed stochastic gradient tracking methods.
\newblock {\em Mathematical Programming\/}~{\em 187\/}(1), 409--457.

\bibitem[\protect\citeauthoryear{Sayed et~al.}{Sayed et~al.}{2014}]{sayed2014adaptation}
Sayed, A.~H. et~al. (2014).
\newblock Adaptation, learning, and optimization over networks.
\newblock {\em Foundations and Trends in Machine Learning\/}~{\em 7\/}(4-5), 311--801.

\bibitem[\protect\citeauthoryear{Spiridonoff, Olshevsky, and Paschalidis}{Spiridonoff et~al.}{2020}]{spiridonoff2020robust}
Spiridonoff, A., A.~Olshevsky, and I.~C. Paschalidis (2020).
\newblock Robust asynchronous stochastic gradient-push: Asymptotically optimal and network-independent performance for strongly convex functions.
\newblock {\em Journal of Machine Learning Research\/}~{\em 21\/}(58), 1--47.

\bibitem[\protect\citeauthoryear{Takezawa, Bao, Niwa, Sato, and Yamada}{Takezawa et~al.}{2023}]{takezawa2023momentum}
Takezawa, Y., H.~Bao, K.~Niwa, R.~Sato, and M.~Yamada (2023).
\newblock Momentum tracking: Momentum acceleration for decentralized deep learning on heterogeneous data.
\newblock {\em Transactions on Machine Learning Research\/}.

\bibitem[\protect\citeauthoryear{Tang, Lian, Yan, Zhang, and Liu}{Tang et~al.}{2018}]{tang2018d}
Tang, H., X.~Lian, M.~Yan, C.~Zhang, and J.~Liu (2018).
\newblock $ d^{2} $: Decentralized training over decentralized data.
\newblock In {\em International Conference on Machine Learning}, pp.\  4848--4856. PMLR.

\bibitem[\protect\citeauthoryear{Vogels, He, Koloskova, Karimireddy, Lin, Stich, and Jaggi}{Vogels et~al.}{2021}]{vogels2021relaysum}
Vogels, T., L.~He, A.~Koloskova, S.~P. Karimireddy, T.~Lin, S.~U. Stich, and M.~Jaggi (2021).
\newblock Relaysum for decentralized deep learning on heterogeneous data.
\newblock {\em Advances in Neural Information Processing Systems\/}~{\em 34}, 28004--28015.

\bibitem[\protect\citeauthoryear{Xin, Pu, Nedi{\'c}, and Khan}{Xin et~al.}{2020}]{xin2020general}
Xin, R., S.~Pu, A.~Nedi{\'c}, and U.~A. Khan (2020).
\newblock A general framework for decentralized optimization with first-order methods.
\newblock {\em Proceedings of the IEEE\/}~{\em 108\/}(11), 1869--1889.

\bibitem[\protect\citeauthoryear{Yu, Jin, and Yang}{Yu et~al.}{2019}]{yu2019linear}
Yu, H., R.~Jin, and S.~Yang (2019).
\newblock On the linear speedup analysis of communication efficient momentum sgd for distributed non-convex optimization.
\newblock In {\em International Conference on Machine Learning}, pp.\  7184--7193. PMLR.

\bibitem[\protect\citeauthoryear{Yuan, Alghunaim, Ying, and Sayed}{Yuan et~al.}{2020}]{yuan2020influence}
Yuan, K., S.~A. Alghunaim, B.~Ying, and A.~H. Sayed (2020).
\newblock On the influence of bias-correction on distributed stochastic optimization.
\newblock {\em IEEE Transactions on Signal Processing\/}~{\em 68}, 4352--4367.

\bibitem[\protect\citeauthoryear{Yuan, Chen, Huang, Zhang, Pan, Xu, and Yin}{Yuan et~al.}{2021}]{yuan2021decentlam}
Yuan, K., Y.~Chen, X.~Huang, Y.~Zhang, P.~Pan, Y.~Xu, and W.~Yin (2021).
\newblock Decentlam: Decentralized momentum sgd for large-batch deep training.
\newblock In {\em Proceedings of the IEEE/CVF International Conference on Computer Vision}, pp.\  3029--3039.

\bibitem[\protect\citeauthoryear{Yuan, Ling, and Yin}{Yuan et~al.}{2016}]{yuan2016convergence}
Yuan, K., Q.~Ling, and W.~Yin (2016).
\newblock On the convergence of decentralized gradient descent.
\newblock {\em SIAM Journal on Optimization\/}~{\em 26\/}(3), 1835--1854.

\bibitem[\protect\citeauthoryear{Yurochkin, Agarwal, Ghosh, Greenewald, Hoang, and Khazaeni}{Yurochkin et~al.}{2019}]{yurochkin2019bayesian}
Yurochkin, M., M.~Agarwal, S.~Ghosh, K.~Greenewald, N.~Hoang, and Y.~Khazaeni (2019).
\newblock Bayesian nonparametric federated learning of neural networks.
\newblock In {\em International Conference on Machine Learning}, pp.\  7252--7261. PMLR.

\bibitem[\protect\citeauthoryear{Zhang and You}{Zhang and You}{2019}]{zhang2019decentralized}
Zhang, J. and K.~You (2019).
\newblock Decentralized stochastic gradient tracking for non-convex empirical risk minimization.
\newblock {\em arXiv preprint arXiv:1909.02712\/}.

\end{thebibliography}

\newpage
\begin{appendices}
\section{Some Basic Properties}
\begin{lemma}\label{lem: bound sum barM}
    Under Assumption 3, the following inequality holds,
    \begin{equation}\label{eq: bound sum barM}
    \sum_{t = 0}^{T-1} \mathbb{E}\left\Vert \bar{\mathbf{m}}^{(t)}\right\Vert^2\leq \sum_{t = 0}^{T-1} \mathbb{E}\left\Vert \nabla\bar{\mathbf{f}}(\mathbf{X}^{(t)})\right\Vert^2 + \frac{(1-\beta)\sigma^2T}n.
    \end{equation}
\end{lemma}
\proof{Proof of Lemma \ref{lem: bound sum barM}:} According to the iterative formula for $\mathbf{M}^{(t)}$, we have
$\bar{\mathbf{m}}^{(t)} = \beta \bar{\mathbf{m}}^{(t-1)} + (1-\beta)\nabla\overline{\mathbf{F}}(\mathbf{X}^{(t)},\bm\Xi^{(t)})$.
Noting that $\bm\Xi^{(t)}$ is independent of $\mathcal{F}^{(t)}$, we can get
\begin{equation}\label{eq: bound iter barM}
\begin{aligned}
    \mathbb{E}\left[ \left\Vert \bar{\mathbf{m}}^{(t)}\right\Vert^2\Big|\mathcal{F}^{(t)}\right]
    \leq & \beta\left\Vert \bar{\mathbf{m}}^{(t-1)}\right\Vert^2 + (1-\beta)\left\Vert \nabla \bar{\mathbf{f}}(\mathbf{X}^{(t)})\right\Vert^2 + \frac{(1-\beta)^2\sigma^2}{n}.
\end{aligned}
\end{equation}
which follows from the Cauchy-Schwarz inequality and Assumption 3. Taking full expectation on both sides and iterating, then summing up both sides from $0$ to $T-1$, the lemma holds.
\endproof

\section{Proof of Lemmas}
\proof{Proof of Lemma \ref{lem: iter for f nonconvex}:}
According to the $L$-smoothness of $f_i$, we have, we have
\begin{equation}\label{eq: iter for f(z) 1}
    \begin{aligned}
        & \mathbb{E}[f(\mathbf{z}^{(t+1)}) |\mathcal{F}^{(t)}] \leq f(\mathbf{z}^{(t)}) + \langle\nabla f(\mathbf{z}^{(t)}),-\alpha\nabla \bar{\mathbf{f}}(\mathbf{X}^{(t)})\rangle + \frac{\alpha^2L}2\left\Vert\nabla\bar{\mathbf{f}}(\mathbf{X}^{(t)})\right\Vert^2 + \frac{\alpha^2L\sigma^2}{2n}.
    \end{aligned}
\end{equation}

\begin{equation}\label{eq: iter for f(z) 1.1}
\begin{aligned}
    & \langle\nabla f(\mathbf{z}^{(t)}),-\alpha\nabla \bar{\mathbf{f}}(\mathbf{X}^{(t)})\rangle\\
    \leq & \frac{1-\beta}{2L\beta}\left\Vert\nabla f(\mathbf{z}^{{(t)}}) - \nabla \bar{\mathbf{f}}(\overline{\mathbf{X}}^{(t)})\right\Vert^2 + \frac{\alpha^2L\beta}{2(1-\beta)}\left\Vert\nabla \bar{\mathbf{f}}(\mathbf{X}^{(t)})\right\Vert^2 - \alpha\langle\nabla \bar{\mathbf{f}}(\overline{\mathbf{X}}^{(t)}),\nabla \bar{\mathbf{f}}(\mathbf{X}^{(t)})\rangle.
\end{aligned}
\end{equation}

The smoothness of $f_i$ and \eqref{eq: def of z} implies $\Vert\nabla f(\mathbf{z}^{(t)}) - \nabla \bar{\mathbf{f}}(\overline{\mathbf{X}}^{(t)})\Vert^2 \leq \frac{\alpha^2L^2\beta^2}{(1-\beta)^2}\Vert \bar{\mathbf{m}}^{(t-1)} \Vert^2$. Using the fact that $2\langle \mathbf{a},\mathbf{b}\rangle = \Vert\mathbf{a}\Vert^2 + \Vert\mathbf{b}\Vert^2 - \Vert\mathbf{a-b}\Vert^2$, we have
\begin{equation}\label{eq: iter for f(z) 1.1(2)}
\begin{aligned}
    & \langle\nabla \bar{\mathbf{f}}(\overline{\mathbf{X}}^{(t)}),\nabla \bar{\mathbf{f}}(\mathbf{X}^{(t)})\rangle
    \geq \frac12\left\Vert\nabla \bar{\mathbf{f}}(\overline{\mathbf{X}}^{(t)})\right\Vert^2 + \frac12\left\Vert\nabla \bar{\mathbf{f}}(\mathbf{X}^{(t)})\right\Vert^2 - \frac{L^2}{2n}\left\Vert\overline{\mathbf{X}}^{(t)} - \mathbf{X}^{(t)}\right\Vert_{\mathrm{F}}^2.
\end{aligned}
\end{equation}
Substituting them into \eqref{eq: iter for f(z) 1} confirms the validity of the lemma.
\endproof

\proof{Proof of Lemma \ref{lem: iter for f PL}:}
    Using $2\langle\mathbf{a},\mathbf{b}\rangle = \left\Vert \mathbf{a}\right\Vert^2 + \left\Vert \mathbf{b}\right\Vert^2 - \left\Vert \mathbf{a}-\mathbf{b}\right\Vert^2$ in \eqref{eq: iter for f(z) 1} and $\left\Vert \nabla f(\mathbf{z}^{(t)})\right\Vert^2 \leq 2\tilde{f}(\mathbf{z}^{(t)})$, we have
    \begin{equation*}
    \begin{aligned}
        & \mathbb{E}[\tilde{f}(\mathbf{z}^{(t+1)})|\mathcal{F}^{(t)}]\\
        \leq & \tilde{f}(\mathbf{z}^{(t)}) - \frac{\alpha}{2}\left\Vert \nabla f(\mathbf{z}^{(t)})\right\Vert^2 - \frac{\alpha(1-\alpha L)}{2}\left\Vert \nabla \bar{\mathbf{f}}(\mathbf{X}^{(t)})\right\Vert^2 + \frac{\alpha}{2}\left\Vert \nabla f(\mathbf{z}^{(t)}) - \nabla \bar{\mathbf{f}}(\mathbf{X}^{(t)})\right\Vert^2 + \frac{\alpha^2L\sigma^2}{2n}\\
        \leq & (1-\alpha\mu)\tilde{f}(\mathbf{z}^{(t)}) - \frac{\alpha(1-\alpha L)}{2}\left\Vert \nabla \bar{\mathbf{f}}(\mathbf{X}^{(t)})\right\Vert^2 + \frac{\alpha L^2}{2n}\sum_{i = 1}^n\left\Vert \mathbf{z}^{(t)} - \mathbf{x}_i^{(t)}\right\Vert^2 + \frac{\alpha^2L\sigma^2}{2n}\\
        \leq & (1-\alpha\mu)\tilde{f}(\mathbf{z}^{(t)}) - \frac{\alpha(1-\alpha L)}{2}\left\Vert \nabla \bar{\mathbf{f}}(\mathbf{X}^{(t)})\right\Vert^2 + \frac{\alpha L^2}{2n}\left\Vert \mathbf{P}_{\mathbf{I}}\mathbf{X}^{(t)}\right\Vert_{\mathrm{F}}^2 + \frac{\alpha^3 L^2\beta^2}{2(1-\beta)^2}\left\Vert \bar{\mathbf{m}}^{(t-1)}\right\Vert^2 + \frac{\alpha^2L\sigma^2}{2n}.
    \end{aligned}
    \end{equation*}Taking full expectation on both sides yields the desired result.
\endproof

\proof{Proof of Lemma \ref{lem: variance bound}:}
Let $\frac{1}{\alpha}\mathbf{U}^\top (\mathbf{X}^{(t)} - \widetilde{\mathbf{X}}^{(t)}) = (\mathbf{a}_1^{(t)},\cdots,\mathbf{a}_n^{(t)})^\top$ and $\bm\Lambda\mathbf{U}^\top(\widetilde{\mathbf{M}}^{(t)} -\mathbf{M}^{(t)}) = (\mathbf{b}_1^{(t)},\cdots,\mathbf{b}_n^{(t)})^\top$. Taking the difference between \eqref{eq: def of x}  and \eqref{eq: def of tildex}, and multiplying both sides by $\mathbf{U}^\top$ we have
\[\mathbf{a}_i^{(t+1)} = \lambda_i\left(2\mathbf{a}_i^{(t)} - \mathbf{a}_i^{(t-1)}\right) + \mathbf{b}_i^{(t)} - \mathbf{b}_i^{(t-1)}.\]
When $\lambda_i = 0$, we have $\mathbf{a}_i = 0$. When $\lambda_i >0$, define $u_i = \sqrt{\lambda_i}\mathrm{e}^{j\theta}$ and $v_i = \sqrt{\lambda_i}\mathrm{e}^{-j\theta}$, where $\theta_i = \arccos(\sqrt{\lambda_i})$ and $j = \sqrt{-1}$. Let $c_i^{(t)} = \frac{u_i^t - v_i^t}{u_i-v_i}$ and $\delta^{(t)} = \mathbf{U}^{\top}(\nabla \mathbf{F}(\mathbf{X}^{(t)},\bm\Xi^{(t)}) - \nabla \mathbf{f}(\mathbf{X}^{(t)})) = (\bm\delta_1^{(t)},\cdots,\bm\delta_n^{(t)})^\top$. Then 
\begin{equation*}
\begin{aligned}
    \mathbf{a}_i^{(t)} = & \sum_{s = 0}^{t-1}\left(\mathbf{b}_i^{(s)} - \mathbf{b}_i^{(s-1)}\right)\frac{u_i^{t-s} - v_i^{t-s}}{u_i-v_i} = \lambda_i(1-\beta)\sum_{r = 0}^{t-1} \sum_{s = r}^{t-1}(c_i^{(t-s)} - c_i^{(t-s-1)})\beta^{s-r} \bm\delta_i^{(r)}\\
    = & \frac{\lambda_i(1-\beta)}{u_i-v_i}\sum_{r = 0}^{t-1}\bigg(\frac{(u_i-1)(u_i^{t-r} - \beta^{t-r})}{u_i-\beta} - \frac{(v_i-1)(v_i^{t-r} - \beta^{t-r})}{v_i-\beta}\bigg)\bm\delta_i^{(r)}\\
    = & \underbrace{\lambda_i(1-\beta)\bm\delta_i^{(t-1)}}_{\Delta^{(t)}_{i,1}} + \underbrace{\frac{\lambda_i(1-\beta)}{u_i-v_i}\sum_{r = 0}^{t-2}\bigg(\frac{(u_i-1)u_i^{t-r}}{u_i-\beta} - \frac{(v_i-1)v_i^{t-r}}{v_i-\beta}\bigg)\bm\delta_i^{(r)}}_{\Delta^{(t)}_{i,2}} + \underbrace{\frac{\lambda_i(1-\beta)^2}{\lambda_i - 2\lambda_i\beta + \beta^2}\sum_{r = 0}^{t-2}\beta^{t-r}\bm\delta_i^{(r)}}_{\Delta^{(t)}_{i,3}},\\
\end{aligned}
\end{equation*}
where the last inequality is because of $\frac{u_i-1}{u_i-\beta} - \frac{v_i-1}{v_i-\beta} = \frac{(1-\beta)(u_i-v_i)}{(u_i-\beta)(v_i-\beta)} = \frac{(1-\beta)(u_i-v_i)}{\lambda_i - 2\lambda_i\beta + \beta^2}$. Since $u_i \neq \beta$ and $v_i\neq \beta$, the sum of above series is reasonable.
Applying the Cauchy inequality, it follows that
\begin{equation}\label{eq: Cauchy bound for a_i^{(t)}}
    \begin{aligned}
        \left\Vert\mathbf{a}_i^{(t)}\right\Vert^2
        \leq & \left\Vert\Delta^{(t)}_{i,1}\right\Vert^2 + 2\left\Vert\Delta^{(t)}_{i,2}\right\Vert^2 + 2\left\Vert\Delta^{(t)}_{i,3}\right\Vert^2.
    \end{aligned}
\end{equation}

Since $\bm\Xi^{(0)},\cdots,\bm\Xi^{(t)}$ are independent, $\mathcal{F}^{(t)}$ is a martingale. We can check that, for any series $\{d_r\}_{r = 0}^\infty$, we have
\begin{equation}\label{eq: formula for martingale2}
    \begin{aligned}
        \mathbb{E}\bigg\Vert\sum_{r = 0}^{t} d_r\bm\delta^{(r)}\bigg\Vert^2 
        = & \sum_{r = 0}^{t} d_r^2\mathbb{E}\Vert\bm\delta^{(r)}\Vert^2.
    \end{aligned}
\end{equation}

The inequality below will be frequently used in the following proof,
\begin{equation}\label{eq: bound for sum delta}
\begin{aligned}
    \sum_{i = 2}^n \mathbb{E}\Vert\bm\delta_i^{(t)}\Vert^2 \leq  n\sigma^2.
\end{aligned}
\end{equation}
For $\Vert\Delta^{(t)}_{i,1}\Vert^2$, applying \eqref{eq: bound for sum delta}, we have $\sum_{i = 2}^n\mathbb{E}\Vert\Delta^{(t)}_{i,1}\Vert^2 \leq \lambda^2(1-\beta)^2 n\sigma^2$.

Applying \eqref{eq: formula for martingale2} to $\mathbb{E}\Vert\Delta^{(t)}_{i,2}\Vert^2$, we have
\begin{equation}\label{eq: bound for Delta2 temp1}
\begin{aligned}
    \mathbb{E}\left\Vert\Delta^{(t)}_{i,2}\right\Vert^2
    = & \lambda_i^2\sum_{r = 0}^{t-2}\eta_i^{(t-r)}\mathbb{E}\Vert\bm\delta_i^{(r)}\Vert^2,
\end{aligned}
\end{equation}
where $\eta^{(t - r)}$ is defined as follows. Noting that $u_i$ is the conjugate of $v_i$, we have 
\begin{equation}\label{eq: coefficient of Delta_2 contraction 1}
\begin{aligned}
    &\eta_i^{(t-r)} := \left(\frac{1-\beta}{u_i-v_i}\left(\frac{(u_i-1)u_i^{t-r}}{u_i-\beta} - \frac{(v_i-1)v_i^{t-r}}{v_i-\beta}\right)\right)^2 \\
    \leq & \frac{4(1-\beta)^2}{\lambda_i-\lambda_i^2}\frac{\vert(u_i-1)(v_i-\beta)\vert^2 \cdot \vert u_i^{t-r}\vert^2}{(\lambda_i - 2\lambda_i\beta + \beta^2)^2} =  4\lambda_i^{t-r-1}(1-\lambda_i)h_1^2(\beta,\lambda_i) + 4\lambda_i^{t-r}h_2^2(\beta,\lambda_i),
\end{aligned}
\end{equation}
where $h_1(\beta,\lambda_i)=\frac{\beta(1-\beta)}{\lambda_i - 2\lambda_i\beta + \beta^2}$ and $h_2(\beta,\lambda_i) = \frac{(1-\beta)^2}{\lambda_i - 2\lambda_i\beta + \beta^2}$. After some simple analysis, we can get for all $\lambda \in (0,1)$ and $\beta \in [0,1)$, $h_1(\beta,\lambda_i) \leq \frac{1}{2\sqrt{\lambda_i(1-\lambda_i)}}$ and $h_2(\beta,\lambda_i) \leq \frac{1}{\lambda_i}$. Then we can get $\eta_i^{(t)} \leq 5\lambda_i^{t-r-2}$.
Substituting it into \eqref{eq: bound for Delta2 temp1} and summing the inequalities over $i$ from $2$ to $n$, we can get $\sum_{i = 2}^n \mathbb{E}\left\Vert\Delta^{(t)}_{i,2}\right\Vert^2 \leq \frac{5\lambda^2n\sigma^2}{1-\lambda}$.

Finally, we consider the bound of $\mathbb{E}[\Vert\Delta^{(t)}_{i,3}\Vert^2]$. Using \eqref{eq: formula for martingale2} again, we have
\begin{equation*}
\begin{aligned}
    \sum_{i = 2}^n\mathbb{E}\left[\left\Vert\Delta^{(t)}_{i,3}\right\Vert^2\right]
     \leq  &\frac{(1-\beta)}{1-\lambda_i}\sum_{i = 2}^n\lambda_i^2h_3^2(\beta,\lambda_i)\mathbb{E}\Vert\bm\delta_i^{(r)}\Vert^2 \leq \frac{\lambda^2(1-\beta)n\sigma^2}{1-\lambda}.\\
\end{aligned}
\end{equation*}
where $h_3(\beta,\lambda) = \frac{\sqrt{1-\lambda}(1-\beta)\beta^2}{\lambda - 2\lambda \beta + \beta^2}$ and the last inequality is because of $h_3(\beta,\lambda) \leq 1$.

Summing \eqref{eq: Cauchy bound for a_i^{(t)}} from $2$ to $n$,  combining the upper bound of $\sum_{i = 2}^n\mathbb{E}\Vert\Delta^{(t)}_{i,k}\Vert^2 $,$k =1,2,3$, and noting that $\mathbb{E} \Vert\mathbf{P}_{\mathbf{I}}(\mathbf{X}^{(t)} - \widetilde{\mathbf{X}}^{(t)})\Vert_{\mathrm{F}}^2 = \alpha^2\sum_{i = 2}^n\mathbb{E}\Vert\mathbf{a}_i^{(t)}\Vert^2$, 
we can get the desired result.
\endproof

\proof{Proof of Lemma \ref{Lem: transformation for x}:}
Define $\widetilde{\mathbf{S}}^{(t)} = (\mathbf{I} - \mathbf{W})^{1/2} \widetilde{\mathbf{Y}}^{(t-1)} + \alpha \mathbf{W} \tilde{m}^{(t)}$. Let $\widetilde{\mathbf{N}}^{(t)}$ be any series that satisfies $\widetilde{\mathbf{N}}^{(0)} = \nabla \mathbf{f}(\overline{\mathbf{X}}^{(0)})$. For $t \geq 1$, we have
\begin{equation}\label{eq: iter for tildexs(x)}
    \begin{aligned}
        \widetilde{\mathbf{X}}^{(t+1)} = & (2\mathbf{W} - \mathbf{I}) \widetilde{\mathbf{X}}^{(t)} - \widetilde{\mathbf{S}}^{(t)} - \alpha \mathbf{W} \widetilde{\mathbf{R}}^{(t)},
    \end{aligned}
\end{equation}
\begin{equation}\label{eq: iter for tildexs(s)}
    \begin{aligned}
        \widetilde{\mathbf{S}}^{(t+1)} = & \widetilde{\mathbf{S}}^{(t)} + (\mathbf{I} - \mathbf{W})\widetilde{\mathbf{X}}^{(t)} + \alpha \mathbf{W} (\widetilde{\mathbf{N}}^{(t+1)} - \widetilde{\mathbf{N}}^{(t)}).
    \end{aligned}
\end{equation}
If we define $\widetilde{\mathbf{Y}}^{(-1)} = \mathbf{0}$, we can check that \eqref{eq: iter for tildexs(x)} and \eqref{eq: iter for tildexs(s)} also hold for $t=0$.

Let $\mathbf{p}_i^{(t)} = [\mathbf{U} \widetilde{\mathbf{X}}^{(t)}]_{i,\cdot}$ and $\mathbf{q}_i^{(t)} = [(\mathbf{I} - \hat{\bm\Lambda})^{-1/2}\hat{\mathbf{U}}^\top\mathbf{S}^{(t)}]_{i-1,\cdot}$, $i \geq 2$; $\mathbf{z}_i^{(t)} = [\mathbf{U}^\top \widetilde{\mathbf{N}}^{(t)}]_{i,\cdot}$ and $\mathbf{r}_i^{(t)} = [\mathbf{U}^\top \widetilde{\mathbf{R}}^{(t)}]_{i,\cdot}$, where $[\mathbf{A}]_{i,\cdot}$ denotes the column vector of the $i$th row of matrix $\mathbf{A}$. Multiplying both sides of \eqref{eq: iter for tildexs(x)} and \eqref{eq: iter for tildexs(s)} by $\mathbf{U}^\top$, and with some transformation, we can get
\begin{equation*}
\begin{aligned}
    \mathbf{p}_i^{(t+1)} &= (2\lambda_i-1) \mathbf{p}_i^{(t)} - \sqrt{1-\lambda_i} \mathbf{q}_i^{(t)} - \alpha \lambda_i \mathbf{r}_i^{(t)},\\
    \mathbf{q}_i^{(t+1)} & = \mathbf{q}_i^{(t)} + \sqrt{1 - \lambda_i} \mathbf{p}_i^{(t)} + \frac{\alpha\lambda_i}{\sqrt{1-\lambda_i}} (\mathbf{z}_i^{(t+1)} - \mathbf{z}_i^{(t)}).
\end{aligned}
\end{equation*}
Let 
\begin{equation*}
    \begin{aligned}
        G_i = &
        \begin{bmatrix}
        2\lambda_i - 1& -\sqrt{1-\lambda_i}\\
        \sqrt{1-\lambda_i} & 1
        \end{bmatrix} 
        = V_i\Gamma_i V_i^{-1},
        \quad V_i  =
        \begin{bmatrix}
        - 1& -1\\
        \sqrt{1-\lambda_i} + j\sqrt{\lambda_i} & \sqrt{1-\lambda_i} - j\sqrt{\lambda_i}
        \end{bmatrix},
    \end{aligned}
\end{equation*}
and $\Gamma_i = \mathrm{diag}\{\lambda_i + \sqrt{\lambda_i^2 - \lambda_i},\lambda_i - \sqrt{\lambda_i^2 - \lambda_i}\}$, where $\mathrm{diag}\{a_1,\cdots,a_n\}$ represents the diagonal matrix with elements $a_1,\cdots,a_n$ on the diagonal. \cite{Unified_D2_yuan2022} proved that, when $0< \lambda_i < 1$, we have $\Vert\Gamma_i\Vert_{\mathrm{op}} \leq \sqrt{\lambda_i}$, $\left\Vert V_i^{-1}\right\Vert_{\mathrm{op}}^2 \leq \frac{1}{\lambda_i}$ and $\left\Vert V_i\right\Vert_{\mathrm{op}}^2 \leq 4$.

Let $\mathbf{E}^{(t)}_i = V_i^{-1}(\mathbf{p}_i^{(t)},\mathbf{q}_i^{(t)})^\top$, $i = 2,\cdots, n$, and $\mathbf{E}^{(t)} = ((\mathbf{E}^{(t)}_2)^\top,\cdots,(\mathbf{E}^{(t)}_n)^\top)^\top$, then 
\begin{equation}\label{eq: iter formula for eik}
\mathbf{E}_i^{(t+1)} = \Gamma_i \mathbf{E}_i^{(t)} + \alpha \lambda_i V_i^{-1}
\begin{bmatrix}
    -\mathbf{r}_i^{(t)}\\
    (\mathbf{z}_i^{(t+1)} - \mathbf{z}_i^{(t)})/\sqrt{1-\lambda_i}
\end{bmatrix}.
\end{equation}
Using the fact that $\left\Vert\mathbf{a} - \mathbf{b}\right\Vert^2 \leq \frac1{\sqrt{\lambda}}\left\Vert\mathbf{a}\right\Vert^2 + \frac{1}{1-\sqrt{\lambda}}\left\Vert\mathbf{b}\right\Vert^2$, and noting that $\left\Vert\mathbf{A} \mathbf{x}\right\Vert^2 \leq \left\Vert\mathbf{A}\right\Vert_{\mathrm{op}}^2\left\Vert\mathbf{x}\right\Vert^2$, we have
\begin{equation}\label{eq: iter for ek temp0}
    \begin{aligned}
        \left\Vert\mathbf{E}_i^{(t+1)}\right\Vert^2 
        \leq & \sqrt{\lambda}\left\Vert \mathbf{E}_i^{(t)}\right\Vert^2 + \frac{\alpha^2 \lambda}{1-\sqrt{\lambda}} \left\Vert\mathbf{r}_i^{(t)}\right\Vert^2 + \frac{\alpha^2 \lambda}{(1-\sqrt{\lambda})(1-\lambda)}\left\Vert\mathbf{z}_i^{(t+1)} - \mathbf{z}_i^{(t)}\right\Vert^2.
    \end{aligned} 
\end{equation}  

The following two inequalities give the upper bounds of $\sum_{i = 2}^n\Vert\mathbf{r}_i^{(t)}\Vert^2$ and $\sum_{i = 2}^n\Vert\mathbf{z}_i^{(t+1)} - \mathbf{z}_i^{(t)}\Vert^2$.
\begin{equation}\label{eq: bound for r^k}
    \sum_{i = 2}^n\left\Vert\mathbf{r}_i^{(t)}\right\Vert^2 = \left\Vert \hat{\mathbf{U}} \widetilde{\mathbf{R}}^{(t)}\right\Vert_{\mathrm{F}}^2 = \left\Vert \mathbf{P}_{\mathbf{I}} \widetilde{\mathbf{R}}^{(t)}\right\Vert_{\mathrm{F}}^2, 
\end{equation}
\begin{equation}\label{eq: bound for z^k}
    \sum_{i = 2}^n\left\Vert\mathbf{z}_i^{(t+1)} - \mathbf{z}_i^{(t)}\right\Vert^2 = \left\Vert \mathbf{P}_{\mathbf{I}}\left(\widetilde{\mathbf{N}}^{(t+1)} - \widetilde{\mathbf{N}}^{(t)}\right)\right\Vert_{\mathrm{F}}^2.
\end{equation}
Summing both sides of the \eqref{eq: iter for ek temp0} according to the subscript from 2 to $n$, and applying \eqref{eq: bound for r^k} and \ref{eq: bound for z^k} leads to \eqref{eq: iter for ek lemma version} holds. And for \eqref{eq: ineq between x and e}, we have $\Vert \mathbf{P}_\mathbf{I} \widetilde{\mathbf{X}}^{(t)}\Vert_{\mathrm{F}}^2 = \sum_{i = 2}^n \Vert V_i\mathbf{E}_i^{(t)}\Vert^2 \leq 4\Vert \mathbf{E}^{(t)}\Vert_{\mathrm{F}}^2.$
Substituting it into (9) and applying Lemma \ref{lem: variance bound}, we obtain the desired result. This completes the proof.
\endproof

\section{Proof of Theorem \ref{thm: non-convex}}
\proof{Proof of Theorem \ref{thm: non-convex}:}
In this part we choose $\widetilde{\mathbf{N}}^{(t)} = (1-\beta)\sum_{i = 0}^{t}\beta^{t-i}\nabla \mathbf{f}(\overline{\mathbf{X}}^{(i)}) + \beta^{t+1}\nabla \mathbf{f}(\overline{\mathbf{X}}^{(0)})$. Noting that $\sum_{r = 0}^{t}(1-\beta)\beta^{t-r} + \beta^{t+1} = 1$ and the convexity of $\left\Vert \cdot\right\Vert_{\mathrm{F}}^2$, we have the following inequality holds from the Jenson's inequality 
\begin{equation}\label{eq: bound for PI(tildem^kp1 - tildem^k)}
\begin{aligned}
    & \left\Vert \mathbf{P}_{\mathbf{I}}\left(\widetilde{\mathbf{N}}^{(t+1)} - \widetilde{\mathbf{N}}^{(t)}\right)\right\Vert_{\mathrm{F}}^2 = \bigg\Vert \sum_{r = 0}^{t}(1-\beta)\beta^{t-r}\mathbf{P}_{\mathbf{I}}\left(\nabla\mathbf{f}(\overline{\mathbf{X}}^{(r+1)})- \nabla\mathbf{f}(\overline{\mathbf{X}}^{(r)})\right)\bigg\Vert_{\mathrm{F}}^2 \\
    \leq & \sum_{r = 0}^{t}(1-\beta)\beta^{t-r}\left\Vert \mathbf{P}_{\mathbf{I}}\left(\nabla\mathbf{f}(\overline{\mathbf{X}}^{(r+1)})- \nabla\mathbf{f}(\overline{\mathbf{X}}^{(r)})\right)\right\Vert_{\mathrm{F}}^2 \leq n\alpha^2 L^2\sum_{r = 0}^{t}(1-\beta)\beta^{t-r}\left\Vert \bar{\mathbf{m}}^{(r)}\right\Vert^2,
\end{aligned}
\end{equation}

Let $\bm\psi^{(t)} = \sum_{r = 0}^{t}(1-\beta)\beta^{t-r}\left\Vert \bar{\mathbf{m}}^{(r)}\right\Vert^2$. Taking \eqref{eq: bound for PI(tildem^kp1 - tildem^k)} into \eqref{eq: iter for ek lemma version}, we can get
\begin{equation}\label{eq: iter for ek}
    \left\Vert \mathbf{E}^{(t+1)}\right\Vert_{\mathrm{F}}^2 \leq \sqrt{\lambda}\left\Vert \mathbf{E}^{(t)}\right\Vert_{\mathrm{F}}^2 + \frac{\alpha^2 \lambda}{1-\sqrt{\lambda}} \left\Vert \mathbf{P}_{\mathbf{I}}\widetilde{\mathbf{R}}^{(t)}\right\Vert_{\mathrm{F}}^2 + \frac{n\alpha^4L^2 \lambda}{(1-\sqrt{\lambda})(1-\lambda)}\bm\psi^{(t)}.
\end{equation}

Next, we consider the iteration bound for $\Vert \mathbf{P}_{\mathbf{I}}\widetilde{\mathbf{R}}^{(t)}\Vert_{\mathrm{F}}^2$,
\begin{equation*}
\begin{aligned}
    \widetilde{\mathbf{R}}^{(t+1)} = \widetilde{\mathbf{M}}^{(t+1)} - \widetilde{\mathbf{N}}^{(t+1)} = \beta\widetilde{\mathbf{R}}^{(t)} + (1-\beta)\left(\nabla \mathbf{f}(\mathbf{X}^{(t+1)}) - \nabla \mathbf{f}(\overline{\mathbf{X}}^{(t)})\right).
\end{aligned}
\end{equation*}
Applying the $L$-smoothness of $\nabla f_i(\mathbf{x})$ and Lemma \ref{Lem: transformation for x}, we have 
\begin{equation}\label{eq: iter for PIwk}
\begin{aligned}
    & \mathbb{E}\left\Vert \mathbf{P}_{\mathbf{I}}\widetilde{\mathbf{R}}^{(t+1)}\right\Vert_{\mathrm{F}}^2 \leq \beta\mathbb{E}\left\Vert \mathbf{P}_{\mathbf{I}}\widetilde{\mathbf{R}}^{(t)}\right\Vert_{\mathrm{F}}^2 + (1-\beta)\mathbb{E}\left\Vert \nabla \mathbf{f}(\mathbf{X}^{(t+1)}) - \nabla \mathbf{f}(\overline{\mathbf{X}}^{(t+1)})\right\Vert_{\mathrm{F}}^2\\
    \leq & \beta\mathbb{E}\left\Vert \mathbf{P}_{\mathbf{I}}\widetilde{\mathbf{R}}^{(t)}\right\Vert_{\mathrm{F}}^2 + (1-\beta)L^2\mathbb{E}\left\Vert \mathbf{X}^{(t+1)} - \overline{\mathbf{X}}^{(t+1)}\right\Vert_{\mathrm{F}}^2\\
    \leq & \beta\mathbb{E}\left\Vert \mathbf{P}_{\mathbf{I}}\widetilde{\mathbf{R}}^{(t)}\right\Vert_{\mathrm{F}}^2 + 8(1-\beta)L^2\mathbb{E}\left\Vert \mathbf{E}^{(t+1)}\right\Vert_{\mathrm{F}}^2 + \frac{26n\alpha^2L^2\lambda^2(1-\beta)\sigma^2}{1-\lambda}.
\end{aligned}
\end{equation}
Let 
\[\mathbf{G}_{\mathbf{E}, \widetilde{\mathbf{R}}} = \begin{bmatrix}
    \sqrt{\lambda} & \frac{\alpha^2\lambda}{1-\sqrt{\lambda}}\\
    8(1-\beta)L^2 & \beta
\end{bmatrix},\quad 
\mathbf{b}_{\mathbf{E}, \widetilde{\mathbf{R}}} = 
\begin{bmatrix}
    \frac{n\alpha^4L^2 \lambda}{(1-\sqrt{\lambda})(1-\lambda)}\\
    0
\end{bmatrix},\quad \tilde{\sigma}^2 = \frac{26n\alpha^2L^2\lambda^2(1-\beta)\sigma^2}{1-\lambda}.\] 
Let $g(x) = \det(\mathbf{G}_{\mathbf{E}, \widetilde{\mathbf{R}}} - x\mathbf{I}) = (x-\sqrt{\lambda})(x-\beta) - \frac{8\alpha^2L^2\lambda(1-\beta)}{1-\sqrt{\lambda}}$. Let $\lambda_{\mathbf{G},i},i = 1,2$, be the solution of $g(x) = 0$. When $\alpha \leq \frac{1-\sqrt{\lambda}}{4L}$, we have
\begin{equation*}
    \begin{aligned}
        g(1) = & (1-\sqrt{\lambda})(1-\beta) - \frac{8\alpha^2L^2\lambda(1-\beta)}{1-\sqrt{\lambda}} > \frac{(1-\sqrt{\lambda})(1-\beta)}{2}.
    \end{aligned}
\end{equation*}
We have $\sqrt{\lambda} + \beta-1 < \lambda_{\mathbf{G},1}< \lambda_{\mathbf{G},2} < 1$, which implies $\rho(\mathbf{G}_{\mathbf{E}, \widetilde{\mathbf{R}}})  < 1$. Then we have
\begin{equation*}
\begin{aligned}
    \begin{bmatrix}
        \mathbb{E}\left\Vert \mathbf{E}^{(t+1)}\right\Vert_{\mathrm{F}}^2 \\
        \mathbb{E}\left\Vert \mathbf{P}_{\mathbf{I}}\widetilde{\mathbf{R}}^{(t+1)}\right\Vert_{\mathrm{F}}^2
    \end{bmatrix} 
    \leq & \mathbf{G}_{\mathbf{E}, \widetilde{\mathbf{R}}} 
    \begin{bmatrix}
        \mathbb{E}\left\Vert \mathbf{E}^{(t)}\right\Vert_{\mathrm{F}}^2 \\
        \mathbb{E}\left\Vert \mathbf{P}_{\mathbf{I}}\widetilde{\mathbf{R}}^{(t)}\right\Vert_{\mathrm{F}}^2
    \end{bmatrix}
    + \mathbb{E}\bm\psi^{(t)}\mathbf{b}_{\mathbf{E}, \widetilde{\mathbf{R}}} + 
    \begin{bmatrix}
        0\\
        \tilde{\sigma}^2
    \end{bmatrix} \\
    \leq & \mathbf{G}_{\mathbf{E}, \widetilde{\mathbf{R}}}^{(t)} 
    \begin{bmatrix}
        \mathbb{E}\left\Vert \mathbf{E}^{(1)}\right\Vert_{\mathrm{F}}^2 \\
        \mathbb{E}\left\Vert \mathbf{P}_{\mathbf{I}}\widetilde{\mathbf{R}}^{(1)}\right\Vert_{\mathrm{F}}^2
    \end{bmatrix}
    + \sum_{s = 1}^t \mathbb{E}\bm\psi^{(s)} \mathbf{G}_{\mathbf{E}, \widetilde{\mathbf{R}}}^{t-s}\mathbf{b}_{\mathbf{E}, \widetilde{\mathbf{R}}} + 
    \sum_{s = 1}^t \mathbf{G}_{\mathbf{E}, \widetilde{\mathbf{R}}}^{t-s}
    \begin{bmatrix}
        0\\
        \tilde{\sigma}^2
    \end{bmatrix}.
\end{aligned}
\end{equation*}
Then 
\begin{equation*}
\begin{aligned}
    \sum_{k = 1}^{T}\begin{bmatrix}
        \mathbb{E}\left\Vert \mathbf{E}^{(t)}\right\Vert_{\mathrm{F}}^2 \\
        \mathbb{E}\left\Vert \mathbf{P}_{\mathbf{I}}\widetilde{\mathbf{R}}^{(t)}\right\Vert_{\mathrm{F}}^2
    \end{bmatrix} 
    \leq & (\mathbf{I} - \mathbf{G}_{\mathbf{E}, \widetilde{\mathbf{R}}})^{-1} \left(
    \begin{bmatrix}
        \mathbb{E}\left\Vert \mathbf{E}^{(1)}\right\Vert_{\mathrm{F}}^2 \\
        \mathbb{E}\left\Vert \mathbf{P}_{\mathbf{I}}\widetilde{\mathbf{R}}^{(1)}\right\Vert_{\mathrm{F}}^2
    \end{bmatrix}
    + \sum_{t = 1}^{T-1}\mathbb{E}\bm\psi^{(t)} \mathbf{b}_{\mathbf{E}, \widetilde{\mathbf{R}}} + 
    \begin{bmatrix}
        0\\
        (T-1)\tilde{\sigma}^2
    \end{bmatrix}\right).
\end{aligned}
\end{equation*}
Noting that $\det(\mathbf{I} - \mathbf{G}_{\mathbf{E}, \widetilde{\mathbf{R}}}) = g(1) \geq \frac{(1-\sqrt{\lambda})(1-\beta)}{2}$, we have
\begin{equation}\label{eq: iter for sum ek temp1}
\begin{aligned}
    &\sum_{t = 1}^{T}\mathbb{E}\left\Vert \mathbf{E}^{(t)}\right\Vert_{\mathrm{F}}^2\leq  \frac{1}{\det(\mathbf{I} - \mathbf{G}_{\mathbf{E}, \widetilde{\mathbf{R}}})}\begin{bmatrix}
        1-\beta & \frac{\alpha^2\lambda}{1-\sqrt{\lambda}}
    \end{bmatrix} 
    \left(
    \begin{bmatrix}
        \mathbb{E}\left\Vert \mathbf{E}^{(1)}\right\Vert_{\mathrm{F}}^2 \\
        \mathbb{E}\left\Vert \mathbf{P}_{\mathbf{I}}\widetilde{\mathbf{R}}^{(1)}\right\Vert_{\mathrm{F}}^2
    \end{bmatrix}
    + \sum_{t = 1}^{T-1}\mathbb{E}\bm\psi^{(t)} \mathbf{b}_{\mathbf{E}, \widetilde{\mathbf{R}}} + 
    \begin{bmatrix}
        0\\
        (T-1)\tilde{\sigma}^2
    \end{bmatrix}\right)\\
    \leq & \begin{bmatrix}
        \frac{2}{1-\sqrt{\lambda}} & \frac{2\alpha^2\lambda}{(1-\sqrt{\lambda})^2(1-\beta)}
    \end{bmatrix} 
    \left(
    \begin{bmatrix}
        \mathbb{E}\left\Vert \mathbf{E}^{(1)}\right\Vert_{\mathrm{F}}^2 \\
        \mathbb{E}\left\Vert \mathbf{P}_{\mathbf{I}}\widetilde{\mathbf{R}}^{(1)}\right\Vert_{\mathrm{F}}^2
    \end{bmatrix}
    + \sum_{t = 1}^{T-1}\mathbb{E}\bm\psi^{(t)} \mathbf{b}_{\mathbf{E}, \widetilde{\mathbf{R}}} + 
    \begin{bmatrix}
        0\\
        (T-1)\tilde{\sigma}^2
    \end{bmatrix}\right)\\
    \leq & \frac{2}{1-\sqrt{\lambda}}\mathbb{E}\left\Vert \mathbf{E}^{(1)}\right\Vert_{\mathrm{F}}^2 + \frac{2n\alpha^4L^2 \lambda}{(1-\sqrt{\lambda})^2(1-\lambda)}\sum_{k = 1}^{T-1}\mathbb{E}\bm\psi^{(k)} +\frac{2\alpha^2\lambda}{(1-\sqrt{\lambda})^2(1-\beta)}\left( \mathbb{E}\left\Vert \mathbf{P}_{\mathbf{I}}\widetilde{\mathbf{R}}^{(1)}\right\Vert_{\mathrm{F}}^2 + (T-1)\tilde{\sigma}^2\right).
\end{aligned}
\end{equation}
Since $\widetilde{\mathbf{R}}^{(0)} = -\beta \nabla\mathbf{f}(\mathbf{X}^{(0)})$, we have $\mathbb{E}\Vert \mathbf{P}_{\mathbf{I}}\widetilde{\mathbf{R}}^{(0)}\Vert_{\mathrm{F}}^2 \leq n\lambda^2\beta^2\zeta_0^2$. Taking it into \eqref{eq: iter for PIwk}, where we choose $k = 0$ we can get 
\begin{equation}\label{eq: iter for bound of w1}
    \mathbb{E}\left\Vert \mathbf{P}_{\mathbf{I}}\widetilde{\mathbf{R}}^{(1)}\right\Vert_{\mathrm{F}}^2\leq n\beta^3\zeta_0^2 + 8(1-\beta)L^2\mathbb{E}\left\Vert \mathbf{E}^{(1)}\right\Vert_{\mathrm{F}}^2 + \tilde{\sigma}^2.
\end{equation}
Next, we consider the bound of $\mathbb{E}\Vert \mathbf{E}^{(1)}\Vert_{\mathrm{F}}^2$. Take the norm of the iterative expression \eqref{eq: iter formula for eik}, and from $\left\Vert \mathbf{A} + \mathbf{B}\right\Vert_{\mathrm{F}}^2 \leq 2\left\Vert \mathbf{A}\right\Vert_{\mathrm{F}}^2 + 2\left\Vert \mathbf{B}\right\Vert_{\mathrm{F}}^2$, we can get 
\begin{equation}\label{eq: iter for bound of e0_i}
    \begin{aligned}
        \left\Vert \mathbf{E}_i^{(1)}\right\Vert^2 \leq & 2\left\Vert\Gamma_i\right\Vert_{\mathrm{op}}^2\left\Vert  \mathbf{E}_i^{(0)}\right\Vert^2 + 2\alpha^2 \lambda_i^2 \left\Vert V_i^{-1}\right\Vert_{\mathrm{op}}^2\left(\left\Vert \mathbf{r}_i^{(0)}\right\Vert^2 + \frac1{1-\lambda_i}\left\Vert \mathbf{z}_i^{(1)} - \mathbf{z}_i^{(0)}\right\Vert^2\right)\\
        \leq & 2\lambda_i \left\Vert V_i^{-1}(\mathbf{p}_i^{(0)},\mathbf{q}_i^{(0)})^\top\right\Vert_{\mathrm{F}}^2 + 4\alpha^2\lambda\left\Vert \mathbf{r}_i^{(0)}\right\Vert^2 + \frac{4\alpha^2\lambda}{1-\lambda}\left\Vert \mathbf{z}_i^{(1)} - \mathbf{z}_i^{(0)}\right\Vert^2\\
        \leq & 8\left\Vert \mathbf{p}_i^{(0)}\right\Vert^2 + 8\left\Vert \mathbf{q}_i^{(0)} \right\Vert^2 + 4\alpha^2\lambda\left\Vert \mathbf{r}_i^{(0)}\right\Vert^2 + \frac{4\alpha^2\lambda}{1-\lambda}\left\Vert \mathbf{z}_i^{(1)} - \mathbf{z}_i^{(0)}\right\Vert^2.
    \end{aligned}
\end{equation}
By the definition of $\mathbf{p}$, $\mathbf{q}$,$\mathbf{r}$ and $\mathbf{z}$, we have
\[\sum_{i = 2}^n\left\Vert \mathbf{p}_i^{(0)}\right\Vert^2 = \left\Vert \widehat{\mathbf{U}}^\top \mathbf{X}^{(0)}\right\Vert^2 = 0,\quad \sum_{i = 2}^n\left\Vert \mathbf{r}_i^{(0)}\right\Vert^2 = \left\Vert \mathbf{P}_\mathbf{I} \widetilde{\mathbf{R}}^{(0)}\right\Vert_{\mathrm{F}}^2 \leq n\beta^2\zeta_0^2,\]
\[\sum_{i = 2}^n\left\Vert \mathbf{q}_i^{(0)}\right\Vert^2 \leq \frac{1}{1-\lambda}\left\Vert \widehat{\mathbf{U}}^\top \mathbf{s}^{(0)}\right\Vert^2 \leq \frac{\alpha^2}{1-\lambda}\left\Vert \mathbf{P}_\mathbf{I} \widetilde{\mathbf{N}}^{(0)}\right\Vert^2 \leq \frac{n\alpha^2(1-\beta)^2 \zeta_0^2}{1-\lambda}.\]
And for the bound of $\sum_{i = 2}^n\Vert \mathbf{z}_i^{(1)}-\mathbf{z}_i^{(0)}\Vert^2= \Vert \mathbf{P}_{\mathbf{I}}(\widetilde{\mathbf{N}}^{(1)} - \widetilde{\mathbf{N}}^{(0)})\Vert_{\mathrm{F}}^2$, we use \eqref{eq: bound for PI(tildem^kp1 - tildem^k)}. Then we have
\begin{equation}\label{eq: iter for bound of e0}
    \begin{aligned}
        \mathbb{E}\left\Vert \mathbf{E}^{(1)}\right\Vert_{\mathrm{F}}^2 = \sum_{i = 2}^n \left\Vert \mathbf{E}_i^{(1)}\right\Vert^2 \leq \frac{8n\alpha^2 (1-\beta)^2 \zeta_0^2}{1-\lambda} + 4n\alpha^2\lambda \beta^2\zeta_0^2 + \frac{4n\alpha^4L^2\lambda\bm\psi^{(0)}}{1-\lambda}.
    \end{aligned}
\end{equation}
Substituting \eqref{eq: iter for bound of w1} and \eqref{eq: iter for bound of e0} into \eqref{eq: iter for sum ek temp1}, we can get
\begin{equation}\label{eq: bound for sum ek}
\begin{aligned}
    & \sum_{t = 1}^{T}\mathbb{E}\left\Vert \mathbf{E}^{(t)}\right\Vert_{\mathrm{F}}^2
    \leq & \frac{n\alpha^2\zeta_0^2}{(1-\sqrt{\lambda})^2}C_0 + \frac{2n\alpha^4L^2\lambda}{(1-\sqrt{\lambda})^2(1-\lambda)}\sum_{t = 0}^{T-1}\mathbb{E}\bm\psi^{(t)} + \frac{52Tn\alpha^4L^2\lambda^3\sigma^2}{(1-\sqrt{\lambda})^2(1-\lambda)},
\end{aligned}
\end{equation}
where $C_0 = \frac{24(1-\beta)^2}{1+\sqrt{\lambda}} + 12\beta^2\lambda(1-\sqrt{\lambda}) + \frac{2\beta^3\lambda}{1-\beta}$.

Next, we consider the upper bound of $\sum_{k = 0}^{T-1}\mathbb{E}\bm\psi^{(k)}$. Based on the definition of $\bm\psi$ and utilizing Lemma \eqref{eq: bound sum barM}, we can obtain
\begin{equation}\label{eq: bound for sum psik}
    \begin{aligned}
        \sum_{k = 0}^{T-1}\mathbb{E}\bm\psi^{(k)} = & \sum_{k = 0}^{T-1}\sum_{s = 0}^{k}(1-\beta)\beta^{s-k}\mathbb{E}\left\Vert \bar{\mathbf{m}}^{(k)}\right\Vert^2\leq \sum_{k = 0}^{T-1}\mathbb{E}\left\Vert \bar{\mathbf{m}}^{(k)}\right\Vert^2\leq \sum_{k = 0}^{T-1}\mathbb{E}\left\Vert \nabla \bar{\mathbf{f}}(\mathbf{X}^{(k)})\right\Vert^2 + \frac{(1-\beta)\sigma^2T}{n}.
    \end{aligned}
\end{equation}
When $\alpha L \leq \frac{1-\beta}{4}$, we can rewrite \eqref{eq: iter for nabla barf} as
\begin{equation}\label{eq: iter for nabla barf v2}
\begin{aligned}
    \mathbb{E}[f(\mathbf{z}^{(t+1)})] \leq & \mathbb{E}[f(\mathbf{z}^{(t)})] +  \frac{\alpha^2L\beta}{2(1-\beta)}\mathbb{E}\left\Vert \bar{\mathbf{m}}^{(t-1)}\right\Vert^2 -\frac{3\alpha}{8}\mathbb{E}\left\Vert \nabla \bar{\mathbf{f}}(\mathbf{X}^{(t)})\right\Vert^2 \\ 
    & + \frac{\alpha^2L\sigma^2}{2n}-\frac{\alpha}2 \mathbb{E}\left\Vert \nabla \bar{\mathbf{f}}(\overline{\mathbf{X}}^{(t)})\right\Vert^2 + \frac{\alpha L^2}{2n}\mathbb{E}\left\Vert \overline{\mathbf{X}}^{(t)} - \mathbf{X}^{(t)}\right\Vert_{\mathrm{F}}^2.
\end{aligned}
\end{equation}
Combining \eqref{eq: bound sum barM}, \eqref{eq: bound for sum ek}, \eqref{eq: bound for sum psik},\eqref{eq: iter for nabla barf v2} and \eqref{eq: iter for ek lemma version}, we have
\begin{equation*}
\begin{aligned}
    & \frac{\alpha}{2}\sum_{t = 0}^{T-1} \left(\frac{1}4\mathbb{E}\left\Vert \nabla \bar{\mathbf{f}}(\mathbf{X}^t)\right\Vert^2 + \mathbb{E}\left\Vert \nabla \bar{\mathbf{f}}(\overline{\mathbf{X}}^t)\right\Vert^2\right)\leq f(\mathbf{x}^{(0)}) - \mathbb{E}[f(\mathbf{z}^{(T)})] + \frac{\alpha^2L\beta}{2(1-\beta)}\sum_{t = 0}^{T-1}\mathbb{E}\left\Vert \bar{\mathbf{m}}^{(t-1)}\right\Vert^2 \\
    &- \frac{\alpha}{4}\sum_{t = 0}^{T-1} \mathbb{E}\left\Vert \nabla \bar{\mathbf{f}}(\mathbf{X}^{(t)})\right\Vert^2 + \frac{\alpha^2L\sigma^2T}{2n} + \frac{4\alpha L^2}{n} \sum_{t = 1}^{T-1}\mathbb{E}\left\Vert \mathbf{E}^{(t)}\right\Vert_{\mathrm{F}}^2 + \frac{13T\alpha^3L^2 \lambda^2\sigma^2}{1-\lambda}\\
    \leq & f(\mathbf{x}^{(0)}) - \mathbb{E}[f(\mathbf{z}^{(T)})] + \frac{\alpha^2L\beta\sigma^2T}{2n} + \frac{\alpha^2L\beta}{2(1-\beta)}\sum_{t = 0}^{T-2}\mathbb{E}\left\Vert \nabla\bar{\mathbf{f}}(\mathbf{X}^{(t)})\right\Vert^2 - \frac{\alpha}{4}\sum_{t = 0}^{T-1} \mathbb{E}\left\Vert \nabla \bar{\mathbf{f}}(\mathbf{X}^{(t)})\right\Vert^2 + \frac{\alpha^2L\sigma^2T}{2n} \\
    & + C_0\frac{4\alpha^3L^2\zeta_0^2}{(1-\sqrt{\lambda})^2} + \frac{8\alpha^5L^4}{(1-\sqrt{\lambda})^2(1-\lambda)}\sum_{k = 0}^{T-1}\mathbb{E}\left\Vert \nabla \bar{\mathbf{f}}(\mathbf{X}^{(k)})\right\Vert^2\\
    & + \frac{8T\alpha^5L^4\lambda(1-\beta)\sigma^2}{n(1-\sqrt{\lambda})^2(1-\lambda)} + \frac{208T\alpha^5L^4\lambda^3(1-\beta)\sigma^2}{(1-\sqrt{\lambda})^2(1-\lambda)} + \frac{13T\alpha^3L^2 \lambda^2\sigma^2}{1-\lambda} \\
    \leq & f(\mathbf{x}^{(0)}) - \mathbb{E}[f(\mathbf{z}^{(T)})] + \frac{\alpha^2L\sigma^2T}{n} + \frac{4C_0\alpha^3L^2\zeta_0^2}{(1-\sqrt{\lambda})^2} + \frac{26 T\alpha^3L^2\lambda^2\sigma^2}{1-\lambda}.
\end{aligned}
\end{equation*}
Dividing both sides of the equation by $\alpha T/2$, we finally obtain the desired result.
\endproof

\section{Proof of Theorem \ref{thm: PL}}
\begin{lemma}\label{lem: Lip lemma}
    For any $L$-smooth function $f$, we have
\[\left\Vert \nabla f(\mathbf{x})\right\Vert^2 \leq 2L(f(\mathbf{x}) - f^\star).\]
\end{lemma}
\proof{Proof of Lemma \ref{lem: Lip lemma}:}
    See \cite{polyak1987introduction}.
\endproof

Next, we prove Theorem \ref{thm: PL}.
\proof{Proof of Theorem \ref{thm: PL}:}
In this part, we choose $\widetilde{\mathbf{N}}^{(t)} = \nabla \mathbf{f}(\bar{\mathbf{X}}^{(t)})$. Let $\{\mathbf{E}^{(t)}\}_{t\geq 1}$ be the series generated from the proof of lemma (10) with $\widetilde{\mathbf{N}}^{(t)} = \nabla \mathbf{f}(\bar{\mathbf{X}}^{(t)})$. Then form assumption \ref{assump of L-smooth}, we have 
\[ \left\Vert \mathbf{P}_{\mathbf{I}}(\widetilde{\mathbf{N}}^{(t+1)} - \widetilde{\mathbf{N}}^{(t)})\right\Vert_{\mathrm{F}}^2 \leq n\alpha^2L^2\left\Vert \bar{\mathbf{m}}^{(t)}\right\Vert^2.\] Substitute it into (14) , we derive that
\begin{equation}\label{eq: iter for ek PLversion}
    \mathbb{E}\left\Vert \mathbf{E}^{(t+1)}\right\Vert_{\mathrm{F}}^2 \leq \sqrt{\lambda}\mathbb{E}\left\Vert \mathbf{E}^{(t)}\right\Vert_{\mathrm{F}}^2 + \frac{\alpha^2 \lambda}{1-\sqrt{\lambda}} \mathbb{E}\left\Vert \mathbf{P}_{\mathbf{I}} \widetilde{\mathbf{R}}^{(t)}\right\Vert_{\mathrm{F}}^2 + \frac{n\alpha^4L^2 \lambda}{(1-\sqrt{\lambda})(1-\lambda)}\mathbb{E}\left\Vert \bar{\mathbf{m}}^{(t)}\right\Vert^2.
\end{equation}

Next we consider the iteration bound for $\mathbb{E}\Vert \mathbf{P}_\mathbf{I} \widetilde{\mathbf{R}}^{(t)}\Vert_{\mathrm{F}}$ under the newly defined $\tilde{\mathbf{m}}^{(t)}$. The iteration formula of $\widetilde{\mathbf{R}}^{(t)}$ can be written as follows
\[\widetilde{\mathbf{R}}^{(t+1)} = \beta\widetilde{\mathbf{R}}^{(t)} + \beta\left(\nabla \mathbf{f}(\bar{\mathbf{X}}^{(t)}) - \nabla \mathbf{f}(\bar{\mathbf{X}}^{(t+1)})\right) + (1-\beta)\left(\nabla \mathbf{f}(\mathbf{X}^{(t+1)}) - \nabla \mathbf{f}(\bar{\mathbf{X}}^{(t+1)})\right).\]
Using Jensen's inequality (the weight of each item is $\beta, \frac{1-\beta}2$ and $\frac{1-\beta}2$), and applying (15), we have
\begin{equation} \label{eq: iter for wk temp1}
\begin{aligned} 
    & \mathbb{E}\left\Vert \mathbf{P}_{\mathbf{I}}\widetilde{\mathbf{R}}^{(t+1)}\right\Vert_{\mathrm{F}}^2
    \leq \beta \mathbb{E}\left\Vert \mathbf{P}_{\mathbf{I}}\widetilde{\mathbf{R}}^{(t)}\right\Vert_{\mathrm{F}}^2 + \frac{2n\alpha^2L^2\beta^2}{1-\beta}\mathbb{E}\left\Vert \bar{\mathbf{m}}^{(t)}\right\Vert^2 + 16(1-\beta)L^2\mathbb{E}\left\Vert \mathbf{E}^{(t+1)}\right\Vert_{\mathrm{F}}^2 + \frac{52n\alpha^2L^2\lambda^2(1-\beta)\sigma^2}{1-\lambda}.
\end{aligned}
\end{equation}
For $t\geq 1$, denote $h^{(t)}_1 = \mathbb{E}[\tilde{f}(\mathbf{z}^{(t)})]$, $h^{(t)}_2 = \mathbb{E}\Vert \mathbf{P}_{\mathbf{I}}\widetilde{\mathbf{R}}^{(t-1)}\Vert_{\mathrm{F}}^2$, $h^{(t)}_3 = \mathbb{E}\Vert \mathbf{E}^{(t)}\Vert_{\mathrm{F}}^2$, and $h^{(t)}_4 = \mathbb{E}\Vert \bar{\mathbf{m}}^{(t-1)}\Vert^2$. Taking them into (7), and applying (15), we have
\begin{equation*}
\begin{aligned}
    h_1^{(t+1)} \leq& (1-\alpha\mu)h_1^{(t)} + \frac{4\alpha L^2}{n}h_3^{(t)} + \frac{\alpha^3 L^2\beta^2}{2(1-\beta)^2}h_4^{(t)} \\
    &- \frac{\alpha(1-\alpha L)}{2}\mathbb{E}\left\Vert \nabla \bar{\mathbf{f}}(\mathbf{X}^{(t)})\right\Vert^2 + \frac{13\alpha^3L^2\lambda^2\sigma^2}{1-\lambda} + \frac{\alpha^2L\sigma^2}{2n}.
\end{aligned}
\end{equation*} 

From \eqref{eq: iter for wk temp1} and \eqref{eq: bound iter barM}, we have
\begin{equation*}
\begin{aligned}
h^{(t+1)}_2\leq \beta h^{(t)}_2 + 16(1-\beta)L^2h_3^{(t)} + \frac{2n\alpha^2L^2\beta^2}{1-\beta}h_4^{(t)} + \left(\frac{10(1-\beta)}{1-\lambda}+3\right)4n\lambda^2\alpha^2L^2\sigma^2,
\end{aligned}
\end{equation*}
\begin{equation}\label{eq: iter for h_4}
    h_4^{(t+1)} \leq \beta  h_4^{(t)} + (1-\beta)\mathbb{E}\left\Vert \nabla \bar{\mathbf{f}}(\mathbf{X}^{(t)})\right\Vert^2 + \frac{(1-\beta)^2\sigma^2}{n}.
\end{equation}

By substituting the inequality into \eqref{eq: iter for ek PLversion}. Under the condition $\alpha L \leq \min\{\frac{1-\sqrt{\lambda}}{10}, \frac{1-\beta}{5}\}$, we have the relation $\sqrt{\lambda} + \frac{16\alpha^2L^2\lambda(1-\beta)}{1-\sqrt{\lambda}}\leq \frac{8+17\sqrt{\lambda}}{25}$. This leads to
\begin{equation}\label{eq: iter for h_3 beta leq lambda}
\begin{aligned}
    h_3^{(t+1)} \leq & \frac{8+17\sqrt{\lambda}}{25}h_3^{(t)} + \frac{\alpha^2 \lambda\beta}{1-\sqrt{\lambda}}h_2^{(t)}  + \frac{n\alpha^3L\lambda\beta}{2(1-\sqrt{\lambda})}h_4^{(t)} \\
    & + \frac{52n\alpha^4L^2\lambda^3(1-\beta)\sigma^2}{(1-\sqrt{\lambda})(1-\lambda)} + \frac{\alpha^4L^2\lambda(1-\beta)^2\sigma^2}{(1-\sqrt{\lambda})(1-\lambda)} + \frac{n\alpha^4L^2\lambda(1-\beta)}{(1-\sqrt{\lambda})(1-\lambda)}\mathbb{E}\left\Vert \nabla\bar{\mathbf{f}}(\mathbf{x}^{(t)})\right\Vert^2.
\end{aligned}
\end{equation}

Denote $\mathcal{L}^{(t)} = h_1^{(t)} + \frac{15\alpha^3L^2\lambda}{n(1-\sqrt{\lambda})^2(1-\beta)}h_2^{(t)} + \frac{12\alpha L^2}{n(1-\sqrt{\lambda})}h_3^{(t)} + \frac{2\alpha^3L^2}{(1-\sqrt{\lambda})(1-\beta)^2}h_4^{(t)}$. Following the proof of theorem 3.8 in \cite{huang2024accelerated}, we can get, when $t \geq 1$,
\begin{equation}\label{eq: bounde for calL}
\begin{aligned}
    \mathcal{L}^{(t+1)} \leq & (1-\alpha \mu)\mathcal{L}^{(t)} + \frac{7\alpha^2L\sigma^2}{10n} + \frac{676\alpha^3L^2\lambda^2\sigma^2}{25(1-\lambda)}.
\end{aligned}
\end{equation}

Next, we consider the condition when $t = 0$, from (7) , $\widetilde{\mathbf{R}}^{(0)} = -\beta \nabla\mathbf{f}(\mathbf{X}^{(0)})$ and $\overline{M}^{(0)} = (1-\beta)\nabla \mathbf{F}(\mathbf{X}^{(0)},\bm\Xi^{(0)})$, we have
\begin{equation}
    h_1^{(1)} = \mathbb{E} \tilde{f}(\mathbf{z}^{(1)}) \leq (1-\alpha\mu)\tilde{f}(\mathbf{x}^{(0)}) - \frac{\alpha(1-\alpha)}{2}\mathbb{E}\left\Vert \nabla \bar{\mathbf{f}} (\mathbf{X}^{(0)})\right\Vert^2 + \frac{\alpha^2L\sigma^2}{2n},
\end{equation}

Noting that $\sum_{i = 2}^n\Vert \mathbf{z}_i^{(1)}-\mathbf{z}_i^{(0)}\Vert^2= \Vert \mathbf{P}_{\mathbf{I}}(\widetilde{\mathbf{N}}^{(1)} - \widetilde{\mathbf{N}}^{(0)})\Vert_{\mathrm{F}}^2 = \Vert \mathbf{P}_{\mathbf{I}}(\nabla \mathbf{f}(\bar{\mathbf{X}}^{(1)}) - \nabla \mathbf{f}(\bar{\mathbf{X}}^{(0)}))\Vert_{\mathrm{F}}^2$, substituting it into \eqref{eq: iter for bound of e0_i}, we can get the upper bound of $h_3^{(1)}$,
\begin{equation}\label{eq: bound for h3}
    \begin{aligned}
        h_3^{(1)} = \sum_{i = 2}^n \left\Vert \mathbf{E}_i^{(1)}\right\Vert^2 \leq \frac{8n\alpha^2 (1-\beta)^2 \zeta_0^2}{1-\lambda} + 4n\alpha^2\lambda \beta^2\zeta_0^2 + \frac{4\alpha^4L^2\lambda(1-\beta)^2\left(n\left\Vert \bar{\mathbf{f}}(\mathbf{X}^{(0)})\right\Vert^2 + \sigma^2\right)}{1-\lambda}.
    \end{aligned}
\end{equation}
And we can check that $h^{(1)}_2 = \mathbb{E}\left\Vert \mathbf{P}_{\mathbf{I}}\widetilde{\mathbf{R}}^{(0)}\right\Vert_{\mathrm{F}}^2\leq n\beta^2\zeta_0^2$ and $h_4^{(1)} \leq (1-\beta)^2\mathbb{E}\left\Vert \nabla\bar{\mathbf{f}}(\mathbf{X}^{(0)})\right\Vert^2 + \frac{(1-\beta)^2\sigma^2}{n}$.

Taking them into the definition of $\mathcal{L}^{(1)}$, we have
\begin{equation}\label{eq: iter for calL1}
\begin{aligned}
    \mathcal{L}^{(1)}
    \leq & (1-\alpha\mu)\left(\tilde{f}(\mathbf{x}^{(0)}) + \frac{D_1\alpha^3L^2\zeta_0^2}{(1-\sqrt{\lambda})^2}\right) + \frac{7\alpha^2L\sigma^2}{10n},
\end{aligned}
\end{equation}
where $D_1 = \frac{50\lambda\beta^2}{3(1-\beta)} + \frac{320(1-\beta)^2}{3(1+\sqrt{\lambda})} + \frac{160}{3}\lambda\beta^2(1-\sqrt{\lambda})$. 

By repeatedly \eqref{eq: bounde for calL} and applying \eqref{eq: iter for calL1}, we ultimately obtain that
\begin{equation}\label{eq: iter for calL}
    \mathcal{L}^{(t)} \leq (1-\alpha \mu)^{t}\left(\tilde{f}(\mathbf{x}^{(0)}) + \frac{D_1\alpha^3L^2\zeta_0^2}{(1-\lambda)^2}\right) + \frac{7\alpha L\sigma^2}{10n\mu} + \frac{676\alpha^2L^2\lambda^2\sigma^2}{25\mu(1-\lambda)}.
\end{equation}

Analogous to the scaling done for $\mathcal{L}^{(t)}$, we can scale $\mathcal{H}^{(t)}$ and $h_4^{(t)}$ following the proof of Theorem 4.2 in \cite{huang2024accelerated}, where we choose $\mathcal{H}^{(t)} =  \frac{5\alpha^2\lambda}{3(1-\sqrt{\lambda})(1-\beta)}h_2^{(t)} + h_3^{(t)} + \frac{2n\alpha^2\lambda}{3(1-\sqrt{\lambda})(1-\beta)}h_4^{(t)}$. For the condition when $t = 0$, we apply \eqref{eq: iter for bound of e0_i} and $\widetilde{\mathbf{N}}^{(t)} = \nabla \mathbf{f}(\overline{\mathbf{X}}^{(t)})$ for a more detailed scaling. Finally, we can obtain that the desired result.
\endproof

\section{Simulations and Real-Data Analysis}
In this section, we demonstrate the advantages of our algorithm through simulations. We compare our algorithm with the following two classes of algorithms: (i) momentum-free bias-correction algorithm (DSGT and ED/D$^2$), and (ii) momentum-based algorithm (DmSGT-HB, DecentLaM, and Quasi-Global). The first type of algorithm is used to show that our algorithm accelerates the convergence of the original one, and the second type of algorithm is used to show that our algorithm has stronger adaptability to data heterogeneity and network sparsity. In the following three parts, we will simulate the square loss function, the general strongly convex objective function, and the non-convex objective function, respectively. 

For the double stochastic communication matrix $\mathbf{W} = (w_{ij})_{1\leq i,j\leq n}$, we model the sparsity of the communication network using a ring graph, where each agent is arranged in a circular formation and can only communicate with its two immediate neighbors. We define the weights as follows: $w_{i,i} = 1/2, w_{i,i+1} = w_{i,i-1} = 1/4$ and $w_{i,j} = 0$ otherwise. In this context, we treat $-1$ as $n$ and $n+1$ as $1$ to account for the cyclic nature of the graph. It can be verified that the spectral gap of the above ring graph satisfies $1-\lambda = \mathcal{O}(1/n^2)$. Consequently, as the number of agents increases, the sparsity of the network significantly increases.

\subsection{Quadratic Loss}
We first consider a simple linear regression problem to visualize the issues caused by network sparsity and data heterogeneity. The loss functions are defined as follows:
\begin{align*}
    f_{i}(\mathbf{x},\bm\xi) = \frac{1}{2}\mathbb{E}_{\mathbf{y}_{i}} \Vert \mathbf{y}_{i} - \mathbf{A}_{i}\mathbf{x} \Vert^2,
\end{align*}
where $i=1,\cdots,n$, $\mathbf{x}\in \mathbb{R}^d,\mathbf{y}_i\in\mathbb{R}^p$, and $\mathbf{A}_{i} = (\mathbf{a}_{i1},\cdots,\mathbf{a}_{ip})^\top \in\mathbb{R}^{p\times d}$ represents the corresponding covariates
at node $i$, with the covariates $\mathbf{a}_{i,j}$ being i.i.d. samples generated from $\mathcal{N}(0,\mathbf{I}_d)$. Here, we choose $d=10$ and $p = 20$. The local optimum parameter $\mathbf{x}_i^\star$ generated as follows: first, let $\mathbf{u}_i$ be i.i.d. samples drawn from $\mathcal{N}(\mathbf{0},\mathbf{I}_d)$. The exact minimizer is given by 
$\mathbf{x}^\star = \left(\sum_{i} \mathbf{A}_{i}^\top\mathbf{A}_{i}\right)^{-1} \left(\sum_{i} \mathbf{A}_{i}^\top\mathbf{A}_{i} \mathbf{u}_i\right)$.
We then define $\mathbf{x}_i^\star = \mathbf{x}^\star + (\mathbf{u}_i - \mathbf{x}^\star)/c$, allowing us to control the data heterogeneity $\zeta^2 = \frac{1}{n}\sum_{i = 1}^n\Vert \nabla f_i(\mathbf{x}^\star)\Vert^2$ through the parameter $c$.
The responses $\mathbf{y}_{i}$ are generated as follows, $\mathbf{y}_{i}^{(t)} = \mathbf{A}_{i} \mathbf{x}_i^\star + \bm\epsilon_{i}^{(t)}$, where $\bm\epsilon_{i}^{(t)}$ are i.i.d. drawn from $ \mathcal{N}(\mathbf{0}_p,\sigma^2\mathbf{I}_p)$. 

\begin{figure*}[ht]
    \centering
    \includegraphics[width=\linewidth]{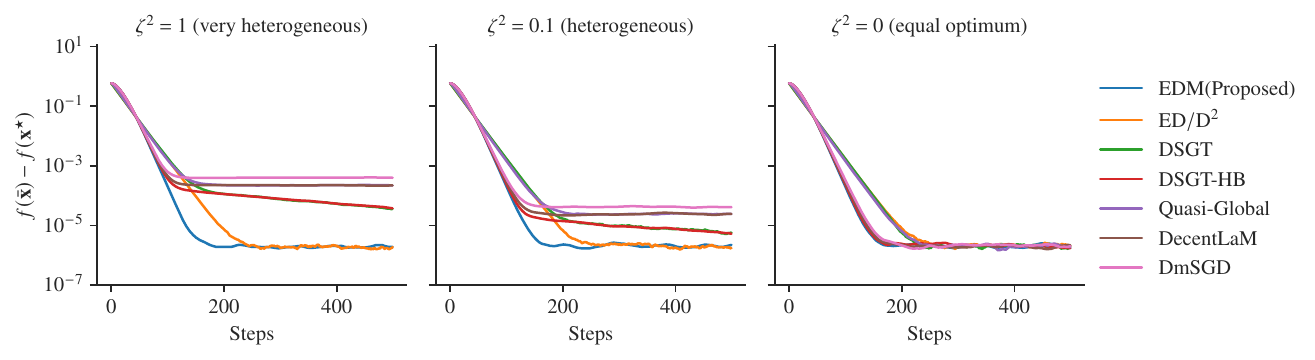}
    \caption{Quadratic loss function. We choose $\sigma^2 = 0.05$, $\alpha = 0.05$, $\lambda = 0.99$ and use different $\zeta^2$ to control the heterogeneity of $f_i$.}
    \label{fig: quadratic loss function}
\end{figure*}

With $n=32$ ($\lambda = 0.99$) and $\sigma^2 = 0.05$, we conducted 20 experiments for various values of $\zeta^2$, and the results are presented in Figure \ref{fig: quadratic loss function}.

The experimental results indicate that all algorithms can converge linearly to a neighborhood of optimal values. However, as data heterogeneity increases, other momentum-based algorithms (DmSGD, Quasi-Global, and DecentLaM) become trapped in heterogeneous regions. Although the DSGT-HB algorithm effectively eliminates heterogeneity errors, its convergence rate is impacted by the original DSGT algorithm, resulting in momentum acceleration failure during the gradient consensus phase. In contrast, the EDM method retains the beneficial properties of the ED/D$^2$, and the introduction of momentum improves the convergence rate.
\subsection{General Strongly Convex Loss}
We consider the $\ell_2$-regularized logistic regression problem to investigate the algorithm's convergence in a general strongly convex case. Let $\mathbf{x}_i \in \mathbb{R}^{d}$ be the local parameter of node $i$, and $\bm\xi_i = \{(\mathbf{u}_{ij}^\top,v_{ij})^\top\}_{j = 1,\cdots,m}$, where $u_{ij}\in \mathbb{R}^d$ is the covariate and $v_{ij}\in\{-1,1\}$ is the response. Given $\mathbf{x}_i$ and $\mathbf{u}_{ij}$, the response $v_{ij}$ is generated such that $v_{ij} = 1$ with probability $1/(1+\exp(-\mathbf{x}_i^\top \mathbf{u}_{ij}))$ and $v_{ij} = -1$ otherwise. The loss function at agent $i$ can be expressed as 
\[f_i(\mathbf{x}_i) = \frac{1}{m}\sum_{j = 1}^m \log[1+\exp(-v_{ij}\mathbf{x}_i^\top \mathbf{u}_{ij})] + \frac{\mu}{2}\left\Vert\mathbf{x}_i\right\Vert^2.\]
It can be verified that this loss function is $\mu$-strongly convex. 

To illustrate the heterogeneity of the data, we consider the following data generation methods. Let $\mathbf{x}_0 = (1,1,\cdots,1)^\top$, and generate $\bm{\epsilon}_i$ i.i.d. from $\mathcal{N}_d(0,\sigma_h^2I_d)$, where $\sigma_h^2$ controls the heterogeneity of local parameter $\mathbf{x}_i$. We define $\mathbf{x}_i = \mathbf{x}_0 + \bm\epsilon_i$. Additionally, the values $\{\mathbf{u}_{ij}\}$ are generated i.i.d. from $\mathcal{N}_d(0,I_d)$. We then generate $z_{ij}$ i.i.d. from uniform distribution $\mathcal{U}(0,1)$. The variable $v_{ij}$ is set to 1 when$z_{ij} \leq 1/(1+\exp(-\mathbf{x}_i^\top \mathbf{u}_{ij}))$, and $-1$ otherwise.

We utilize the full batch gradient $\nabla F(\mathbf{x}_i,\bm\xi_i)$ at every iteration. However, since we are using full batch samples, this results in no randomness. Thus, we add an additional i.i.d. noise term $\mathbf{s}_i^{(t)}$ drawn from $\mathcal{N}(0,\sigma_s^2I_d)$ to $\nabla F(\mathbf{x}_i^{(t)},\bm\xi_i)$. And use $\nabla F(\mathbf{x}_i^{(t)},\bm\xi_i) + \mathbf{s}_i^{(t)}$ as the stochastic gradient. This method is similar to the one used in \cite{Unified_D2_yuan2022} to control the gradient noise.

\begin{figure*}[ht]
    \centering
    \includegraphics[width=\linewidth]{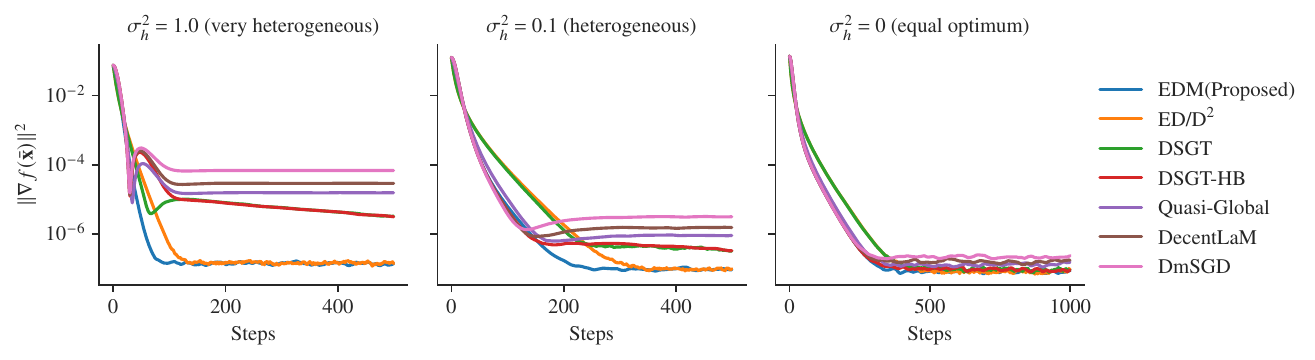}
    \caption{Logistic regression with $l_2$-regularization, random noise version. We choose $\sigma_s^2 = 0.01$, $\alpha = 0.5$, $\lambda = 0.99$, and use $\sigma_h^2$ to control the variance of $\mathbf{x}_i^\star$, which reflects the heterogeneity.}
    \label{fig: Logistic regression with squared penalty}
\end{figure*}

For our simulations, we set $d = 20$ and $m = 2000$, using the $\ell_2$-norm of the gradient $\Vert\nabla \bar{\mathbf{f}}(\overline{\mathbf{X}}^{(t)})\Vert_2^2$ as a measure of convergence. The simulation is conducted 20 times on a ring graph with $n = 32$ nodes. We compare our method against DSGT, ED/D$^2$, DSGT-HB, and DmSGD. The results are presented in Figure \ref{fig: Logistic regression with squared penalty}.

The simulation results demonstrate that our method enhances the performance of ED/D$^2$, achieving convergence to a narrower region of global optimal values while experiencing minimal influence from data heterogeneity.

\subsection{Non-Convex Loss}
Finally, we consider the non-convex condition. We use VGG-11 as the backbone network to train the classification task of the CIFAR10 dataset consisting of $N=50,000$ images of 10 labels, such as cats, airplanes, etc. The photos are $32\times 32$ size with three color channels. The criterion is the cross entropy. To quantitatively analyze the heterogeneity of the data in this dataset, we introduce the Dirichlet distribution, which is widely used to model heterogeneous scenarios for classification problems \citep{yurochkin2019bayesian,lin2021quasi}. We generate $\mathbf{p}_{k} = (p_{k1},\cdots,p_{kn})$ for $ k = 1,\cdots,10$ from $\mathrm{Dirichlet}(\phi,\cdots,\phi)$, allocating $p_{ki}$ proportion of samples with label $k$ to agent $i$. The Dirichlet parameter $\phi$ is used to control the weights assigned to different categories of data across devices, serving as a measure of data heterogeneity. Specifically, a smaller value of $\phi$ indicates greater heterogeneity among the samples allocated to different agents. Our simulation for this part is primarily based on the code provided in \cite{vogels2021relaysum}.

\begin{figure*}[ht]
    \centering
    \includegraphics[width=\linewidth]{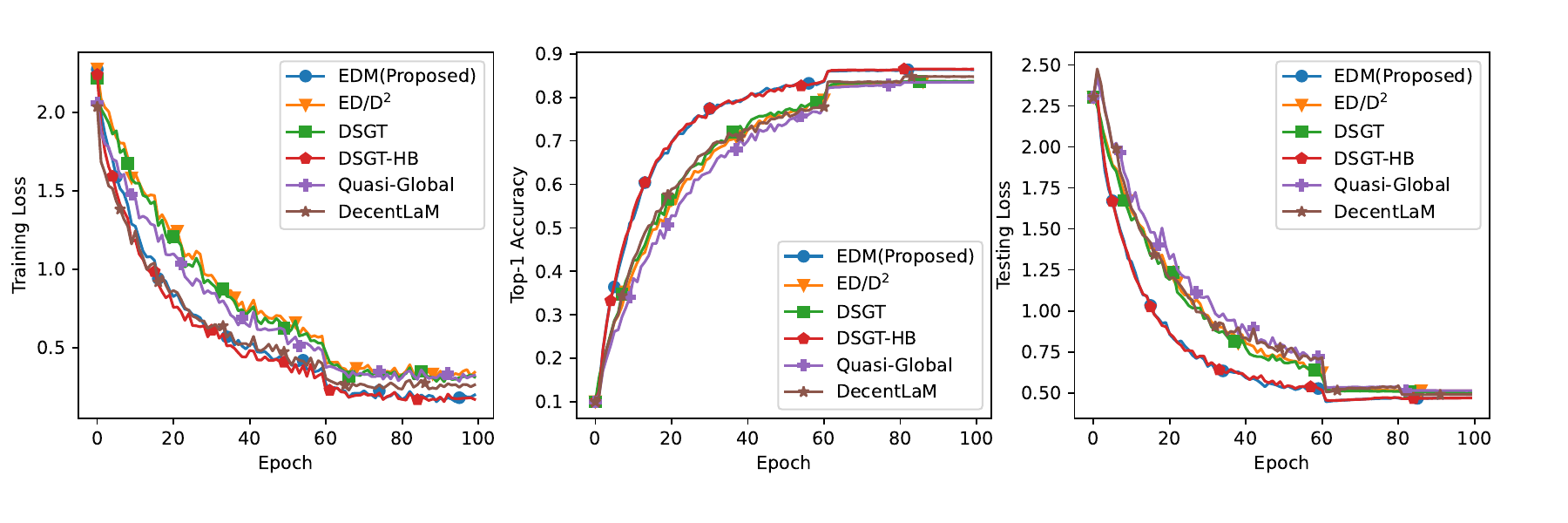}
    \caption{Classification of CIFAR10 by VGG11. Learning rate $\alpha = 0.1$, heterogeneity parameter $\phi = 1.0$ (heterogeneous).}
    \label{fig: CIFAR_alpha1}
\end{figure*}

\begin{figure*}[ht]
    \centering
    \includegraphics[width=\linewidth]{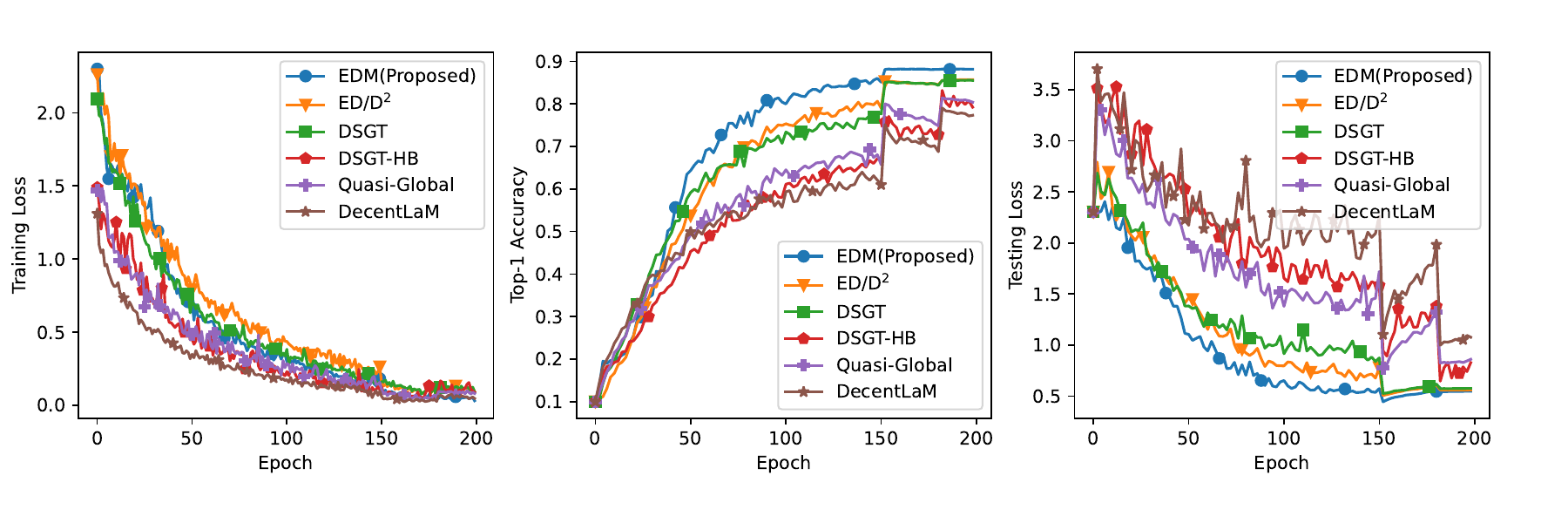}
    \caption{Classification of CIFAR10 by VGG11. Learning rate $\alpha = 0.1$, heterogeneity parameter $\phi = 0.1$ (very heterogeneous).}
    \label{fig: CIFAR_alpha0.1}
\end{figure*}

We considered two cases with $\phi = 1$ (heterogeneous) and $\phi = 0.1$ (highly heterogeneous). For $\phi = 1$, the learning rate is reduced to 10\% of its original value during the 60th and 80th epochs. In the case of $\phi = 0.1$, the learning rate is similarly reduced to 10\% of its original value at the 150th and 180th epochs. Each method is repeated three times, and the results are presented in Figures \ref{fig: CIFAR_alpha1} and \ref{fig: CIFAR_alpha0.1}. Here the training loss is calculated as $\sum_{i = 1}^n f_i(\mathbf{x}_i^{(t)})$. It is important to note that this selection reflects not the convergence properties of the algorithm, but rather whether the algorithm has fallen into the local minimum point for each agent $\mathbf{x}^\star_i$.  In contrast, the testing loss provides a better indication of the convergence performance.

From the simulation results above, we can see that our algorithm outperforms other methods in the classification of CIFAR-10. Specifically, our proposed algorithm exceeds the performance of those that do not utilize momentum acceleration. When $\phi = 1$, both our algorithm and DSGT-HB exhibit comparable convergence performance. However, as heterogeneity increases further ($\phi = 0.1$), the convergence performance of DSGT-HB deteriorates significantly, even becoming inferior to that of the original algorithm. In contrast, our algorithm continues to demonstrate robust convergence properties. These findings are consistent with our theoretical analysis.

\end{appendices}

\end{document}